\documentclass[lettersize,journal]{IEEEtran}
\usepackage{color,soul} 
\soulregister\cite7 
\soulregister\citep7 
\soulregister\citet7 
\soulregister\ref7 
\soulregister\pageref7 
\usepackage{amsmath,amsfonts}
\usepackage{array}
\usepackage{booktabs}
\usepackage[caption=false,font=normalsize,labelfont=sf,textfont=sf]{subfig}
\usepackage{textcomp}
\usepackage{stfloats}
\usepackage{verbatim}
\usepackage{graphicx}
\usepackage{cite}
\usepackage{multirow}
\usepackage[pagewise]{lineno}
\usepackage[table,xcdraw]{xcolor}
\usepackage[numbers]{natbib}
\usepackage{float}

\usepackage{amsthm}
\usepackage{amsmath}
\usepackage{multirow}
\usepackage{soul}
\usepackage{svg}
\usepackage{algorithmic}
\usepackage{makecell}

\usepackage{threeparttable}
\usepackage[linesnumbered,ruled]{algorithm2e}
\usepackage{tikz}

\usepackage{amssymb}
\usepackage{bm}
\usepackage{balance}
\usepackage{url}
\usepackage[colorlinks=true,
            citecolor=blue,
            linkcolor=blue,
            anchorcolor=blue,
            urlcolor=blue]{hyperref}

\usepackage[linesnumbered,ruled]{algorithm2e}

\begin{document}


\title{ME-CPT: Multi-Task Enhanced Cross-Temporal Point Transformer for Urban 3D Change Detection}

\author{
\IEEEauthorblockN{
Luqi Zhang,
Haiping Wang,
Chong Liu,
Zhen Dong,
Bisheng Yang
}
\vspace{-20pt}
\thanks{
This study was jointly supported by the National Natural Science Foundation Project (No. 42130105) and the National Key Research and Development Program of China (No.2022YFB3904105). (Corresponding authors: Zhen Dong and Bisheng Yang)

L. Zhang, H. Wang, C. Liu, Z. Dong, B. Yang are with State Key Laboratory of Information Engineering in Surveying, Mapping and Remote Sensing, Wuhan University, Wuhan 430079, China. (e-mail: lqzhang@whu.edu.cn; hpwang@whu.edu.cn; liuchongwhu@whu.edu.cn; dongzhenwhu@whu.edu.cn; bshyang@whu.edu.cn)
}

}

\markboth{Journal of \LaTeX\ Class Files, ~Vol.~xxx, No.~xxx, xxx~xxx}%
{Shell \MakeLowercase{\textit{et al.}}:ME-CPT: Multi-Task Enhanced Cross-Temporal Point Transformer for Urban 3D Change Detection}

\maketitle

\begin{abstract}

The point clouds collected by the Airborne Laser Scanning (ALS) system provide accurate 3D information of urban land covers. By utilizing multi-temporal ALS point clouds, semantic changes in urban area can be captured, demonstrating significant potential in urban planning, emergency management, and infrastructure maintenance.
Existing 3D change detection methods struggle to efficiently extract multi-class semantic information and change features, still facing the following challenges:
(1) the difficulty of accurately modeling cross-temporal point clouds spatial relationships for effective change feature extraction; (2) class imbalance of change samples which hinders distinguishability of semantic features; (3) the lack of real-world datasets for 3D semantic change detection.
To resolve these challenges, we propose the Multi-task Enhanced Cross-temporal Point Transformer (ME-CPT) network. ME-CPT establishes spatiotemporal correspondences between point cloud across different epochs and employs attention mechanisms to jointly extract semantic change features, facilitating information exchange and change comparison. Additionally, we incorporate a semantic segmentation task and through the multi-task training strategy, further enhance the distinguishability of semantic features, reducing the impact of class imbalance in change types.
Moreover, we release a 22.5 $km^2$ 3D semantic change detection dataset, offering diverse scenes for comprehensive evaluation. Experiments on multiple datasets show that the proposed MT-CPT achieves superior performance compared to existing state-of-the-art methods. The source code and dataset will be released upon acceptance at \url{https://github.com/zhangluqi0209/ME-CPT}.

\end{abstract}

\begin{IEEEkeywords}
Airborne Laser Scanning (ALS); LiDAR; 3D Change Detection;  Cross-Temporal Point Transformer;  Multi-Task Training
\end{IEEEkeywords}

\section{Introduction}
Remote sensing data has been widely used to observe changes occurring on the Earth's surface  over time, with applications spanning environmental monitoring, agricultural surveys, disaster assessment, and map updates \citep{shi_change_2020}.
With rapid urbanization, traditional 2D Change Detection (CD) methods are no longer sufficient to meet growing demands. There is an urgent need for automatic and accurate 3D CD methods to support urban planning, emergency response, and infrastructure development. With the development and increasing diversity of 3D data acquisition technologies, such as laser scanning and oblique photogrammetry, point clouds have become a crucial data source \citep{yang2024ubiquitous}. Compared to traditional remote sensing images, the 3D point cloud offers more intricate geometric details and precise spatial structures, unaffected by challenges such as varying lighting conditions and occlusions \citep{jiang_change_2024}. Using 3D point clouds  enables more comprehensive outcomes, supporting high resolution geometric and attribute for CD in 3D space \citep{stilla_change_2023}.

In recent years, 3D point cloud-based methods for CD have been widely studied and applied in various domains \citep{xiao_3d_2023}. Point cloud data excels at capturing subtle changes in urban terrain, volume, and structures across various types of land cover, enabling applications such as digital twins and urban renewal. Existing 3D CD methods can be broadly classified into two categories: hand-crafted-based methods and deep learning-based methods.

\textbf{Hand-crafted-based methods} are traditional approaches that utilize manually designed features and predefined rules to detect changes \citep{liu_3d_2021}. These methods extract features such as geometric properties, texture characteristics, or statistical attributes from 3D data and compare them to identify areas or objects where changes have occurred. However, the diverse types of land cover and complex dynamic changes in urban environments pose significant challenges to features design. The reliance on empirical thresholds makes them susceptible to false positives or misses. Furthermore, these approaches typically focus on change types of specific objects, limiting their ability to detect diverse semantic changes \citep{tamort20243d}. 

\textbf{Deep learning-based methods} utilize deep neural networks to automatically extract features and patterns from  3D point clouds to detect changes between different epochs. These approaches leverage the capabilities of the deep learning network to extract hierarchical representations directly from raw 3D data, eliminating the reliance on manual feature engineering. Despite their advantages, these methods face three significant challenges:

\begin{enumerate}

\item The limited change features are attributed to the challenge of comprehensively modeling spatial correspondences between multi-temporal point clouds. The extraction of changes in existing work \citep{de_gelis_siamese_2023,wang_end--end_2023} is typically performed by matching points with their nearest or k-nearest neighbors between different epochs and calculating the change feature by subtraction or concatenation. However, the complexity of the spatial distribution and correspondences of cross-temporal point clouds are often overlooked, limiting the model's ability to extract more comprehensive change features.

\item The imbalance of change samples leads to insufficient semantic features representation for multi-temporal points. The imbalance between changed and unchanged samples, as well as between different categories of changes (newly built, demolition, and new clutter) introduces significant training difficulties \citep{zhang2023global}. Change regions typically account for a small proportion of training data, causing unchanged regions to dominate the network feature learning and detection results \citep{8533433}. This imbalance affects feature learning and decision boundaries \citep{zhong2023understanding}, reducing the accuracy and reliability of detection in multiple types of land cover.

\item The lack of available and diverse datasets hinders the development of 3D change detection algorithms\citep{xiao20233d}. Without sufficient datasets covering various change categories, 3D change detection models struggle to learn robust and transferable features, limiting their generalization ability for multiple urban scenarios and making them difficult to apply in real-world applications.

\end{enumerate}


To address the key gaps in existing methods and meet the growing demand for urban 3D semantic change detection, we introduce \underline{M}ulti-Task \underline{E}nhanced \underline{C}ross-Temporal \underline{P}oint \underline{T}ransformer (ME-CPT), an innovative approach for urban 3D semantic change detection. By integrating cross-temporal attention mechanism and multi-task training strategies. The main contributions of the proposed method are summarized as follows:

\begin{enumerate}
    
    \item A novel cross-temporal point transformer is proposed to enhance change feature extraction. By employing a cross-temporal point cloud serialization process, accurate spatiotemporal distribution correspondences are obtained. Additionally, a cross-temporal attention mechanism is utilized to facilitate feature interaction between multi-temporal points, effectively extracting change features.

    \item A multi-task enhancement strategy is incorporated to enhance semantic feature discriminability. This strategy addresses the challenge of insufficient feature extraction caused by the imbalance of change samples. By adopting a multi-task learning strategy, the model enhances feature representation, enabling robust semantic feature extraction.
    
    \item To overcome the lack of large-scale urban scene 3D change detection datasets, we released a novel 3D semantic change detection dataset, specifically designed to tackle the challenges posed by complex urban environments. Spanning a total area of 22.5 $km^2$, the dataset includes diverse scenes and a wide variety of change samples, providing a solid benchmark to evaluate algorithm performance.
\end{enumerate}

The remainder of the paper is organized as follows. Section \ref{sec_related_work} reviews the related works on 3D CD methods. Section \ref{sec_method} elaborates on the proposed network. Section \ref{sec_dataset} introduces the new public 3D semantic change detection dataset. Section \ref{sec_experiment} presents the datasets and quantitative evaluation. Section \ref{sec_discussion} discusses the effectiveness and generalization tests. The conclusion is outlined in Section \ref{sec_conclusion}.

\section{Related Works \label{sec_related_work}}
\subsection{Hand-crafted-based 3D CD methods}
3D CD methods have been extensively studied and applied in various fields such as environmental monitoring, cultural heritage preservation, and urban management. 
Traditional 3D CD methods rely on hand-crafted features or rules, which can be categorized into geometric differencing-based methods, rule-based methods, and traditional machine learning-based algorithms.

\textbf{Geometric differencing based 3D CD methods:}
These methods compare geometric properties like point coordinates, surface normals, object shapes, or volumes between multi-temporal 3D data. For example, the simplest approach calculates point-to-point distances \citep{girardeau2005change}, but it is sensitive to noise and varying point densities. To address these issues, \citet{lague_accurate_2013-1} uses local point cloud normals for surface distance calculation, while \citet{liu_3d_2021} adapts thresholds using k-nearest neighbors distance and local density. These methods are often used in multi-temporal surface analysis, such as depth of erosion of landslides \citep{ma20253d}, changes in rockfall volume \citep{wollenberg-barron_integrating_2024}, and tunnel deformation \citep{pu2024deformation}. However, they are vulnerable to misalignment and noise, leading to inaccurate results, and do not capture semantic changes.

\textbf{Rule-based 3D CD methods:} 
These methods use geometric, textural, or semantic criteria to detect changes between multi-temporal 3D data, typically through rule-based thresholds. Volume, height, and spectral information from different epochs can be used as rules \citep{pang_building_2018, awrangjeb_effective_2018-1}. For example, \citet{dong2018automated} uses ALS point clouds to extract building footprints and features, classifying buildings based on volume change thresholds into newly constructed, height increased or demolished categories. Similarly, \citet{tamort20243d} proposes a semi-automatic method based on ALS point clouds, using rules to identify additions or removals. Although rule-based methods are simple and interpretable, these methods struggle with complex changes and manually defining rules for large datasets can be time-consuming.

\textbf{Machine-learning based 3D CD methods:}
These algorithms commonly detect changes using different classifiers. For instance, \citet{tran_integrated_2018} designs features based on point distributions within multi-temporal point cloud neighborhoods, then uses random forests to classify change categories. \citet{pushkar2018automated} applies Support Vector Machine(SVM) to classify erroneous and building points in construction monitoring data and compare them with existing models to detect changes. Similarly, \citet{peng_building_2016} combines point cloud data with orthophotos, extracting the Digital Surface Model(DSM) and geometric features, and then uses a decision tree to classify building changes. These methods offer a data-driven approach but require extensive feature engineering. This can lead to errors when handling complex scenes or large changes, increasing the risk of inaccuracies.

In summary, although these methods have been widely used, they are often labor intensive and may struggle to capture complex patterns or variations in the data. They often only detect changes for specific objects, limiting their ability to capture the full range of variations in complex scenes. 
\begin{figure*}[h]
    \centering
    \includegraphics[width=0.9\textwidth]{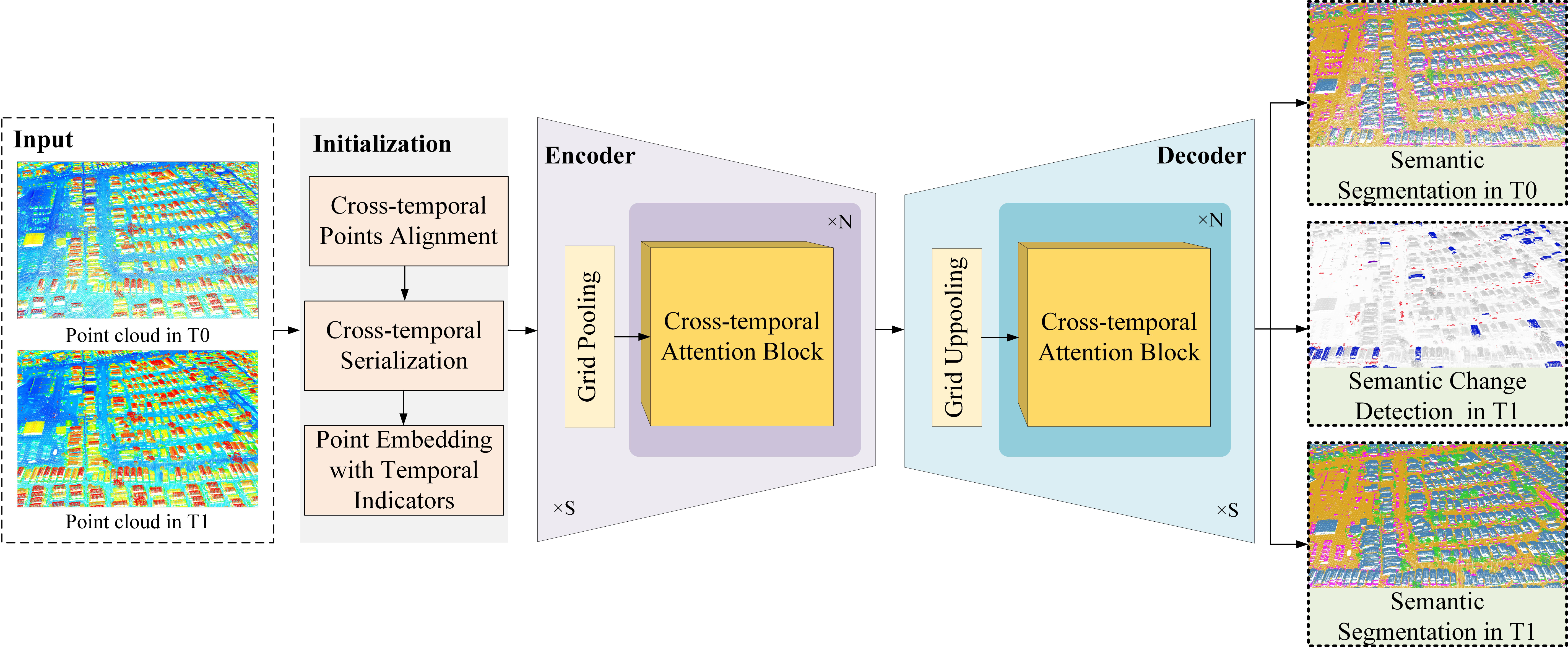}
    \caption{Workflow of the proposed \underline{M}ulti-Task \underline{E}nhanced \underline{C}ross-Temporal \underline{P}oint \underline{T}ransformer (ME-CPT) for urban 3D semantic change detection.}
    \label{fig_workflow}
\end{figure*}

\subsection{Deep-learning based 3D CD methods}
Deep learning techniques have made significant advances in processing 3D point cloud data, owing to their ability to extract high-level features and handle unstructured data \citep{ding_recent_2023}. Deep learning techniques have shown success in automatically detecting changes in satellite images \citep{kaur_developments_2024}.
Inspired by 2D image CD methods, some researchers have transformed 3D point clouds into 2D images for change detection. 
For instance, \citet{zhang_change_2019} converts multi-modal point clouds into images and inputs them into a feed-forward Convolutional Neural Network (CNN) alongside orthophotos, followed by connectivity analysis to obtain the final change detection results.

With the advancements in deep learning techniques for handling irregular 3D point cloud, point-based deep learning change detection methods have gained increasing attention. These methods offer a more direct approach to processing raw 3D data. For example, \citet{krawciw_change_2023} implements binary change detection on a mobile robot platform by generating depth images from both the original point cloud map and the newly scanned point cloud as input, ultimately detecting changes and occlusions in the scene.
\citet{stathoulopoulos2023irregular} uses a deep learning framework to extract features from 3D scan points, then performs voxel-to-point comparisons to detect dynamic changes and identify potential obstacles.

In addition to predicting binary changes, urban environments require change detection results with multi-class semantic information, which is essential for understanding various types of change in complex urban scenes.
\citet{fang_semantic_2023} proposes a semantic support-based change monitoring method that first introduces a semantic segmentation network to extract semantic information from point clouds at different epochs, followed by obtaining results through the comparison of geometric and semantic information. However, this method does not constitute an end-to-end change detection framework. 

Consequently, many researchers have begun to explore end-to-end change detection frameworks that do not require post-processing. For example, \citet{de_gelis_siamese_2023} introduces a Siamese Kernel Point Convolution (KPConv) network for multi-class semantic change detection, utilizing the feature difference obtained from nearest neighboring points. Also, \citet{wang_end--end_2023} achieves point-level change detection in street scenes by employing a local feature aggregation module to extract features, similarly obtaining the nearest feature difference. In addition, \citet{zhan_pgn3dcd_2024} implements a prior-knowledge-guided 3D change detection network by generating prior knowledge for 3D change detection.
In addition to supervised methods, some researchers have explored the application of unsupervised methods for 3D change detection. \citet{de_gelis_dc3dcd_2023-1} proposes an unsupervised method that facilitates point-level multi-class change detection by achieving k-class pseudo-clustering, where users select the corresponding true labels for each predicted cluster. Although the unsupervised method reduces the need for labeled samples, the accuracy in change detection often fails to meet the requirements.

In summary, deep learning-based methods can improve the accuracy and efficiency of change detection tasks in complex 3D urban environments. However, these methods still face challenges in real-world ALS point cloud data, including low accuracy in multi-class semantic change detection, difficulties in data annotation, and imbalanced change samples.

\subsection{Multi-tasks learning mechanism}
The multi-task mechanism is a powerful approach in deep learning that enables models to efficiently learn from related tasks, improving overall performance and generalization while reducing the need for extensive training data for each individual task. 

The multi-task mechanism is widely used in the field of remote sensing image CD.
The multi-task approach for semantic change detection employs a unified framework that enables the simultaneous training of change detection and related tasks, such as segmentation and classification in each epoch. By leveraging multi-task learning, the model can share representations across tasks, allowing it to learn high-level features from diverse supervised labels concurrently. 
Typically, many researchers divide semantic change detection into two modules: semantic segmentation and change localization. Many researchers use multi-temporal remote sensing images as input to separately predict semantic maps and change maps for the two epochs \citep{zuo_multi-task_2024,niu_smnet_2023, cui_mtscd-net_2023}. For example, \citet{zheng_changemask_2022-1} employs a deep multi-task encoder-transformer-decoder framework, which takes into account semantic change relationships and temporal consistency. The network ultimately predicts semantic changes by inheriting the tasks of semantic segmentation and binary change detection. The multi-task mechanism enhance the network's ability to learn and represent features by leveraging the auxiliary task. \citet{shen_semantic_2022} uses building segmentation as a semantic constraint branch to improve the accuracy and completeness of the change detection task. 
In addition to the semantic segmentation task, specific feature enhancement related tasks can also improve the effectiveness of feature learning in the network. \citet{li_detecting_2023} employs building boundary segmentation as an auxiliary task to improve the geometric boundary accuracy of building change detection.

Additionally, the multi-task mechanism has also been applied in deep network frameworks for 3D data processing. For instance, \citet{liu2024transformer} achieves cross-modal change detection by leveraging consistency constraints between the height prediction and semantic segmentation tasks. Beyond change detection, \citet{milli_multi-modal_2023} utilizes a multi-task structure with RGB and LiDAR data to improve road segmentation results. \citet{liebel_generalized_2020} employees DSM data to separately predict roof classification and DSM values, thus refining the DSM data.

In summary, the multi-task mechanism in change detection aims to leverage the synergy between different related tasks to enhance the model's performance, efficiency, and generalization ability.

\begin{figure*}[h]
    \centering
    \includegraphics[width=0.7\textwidth]{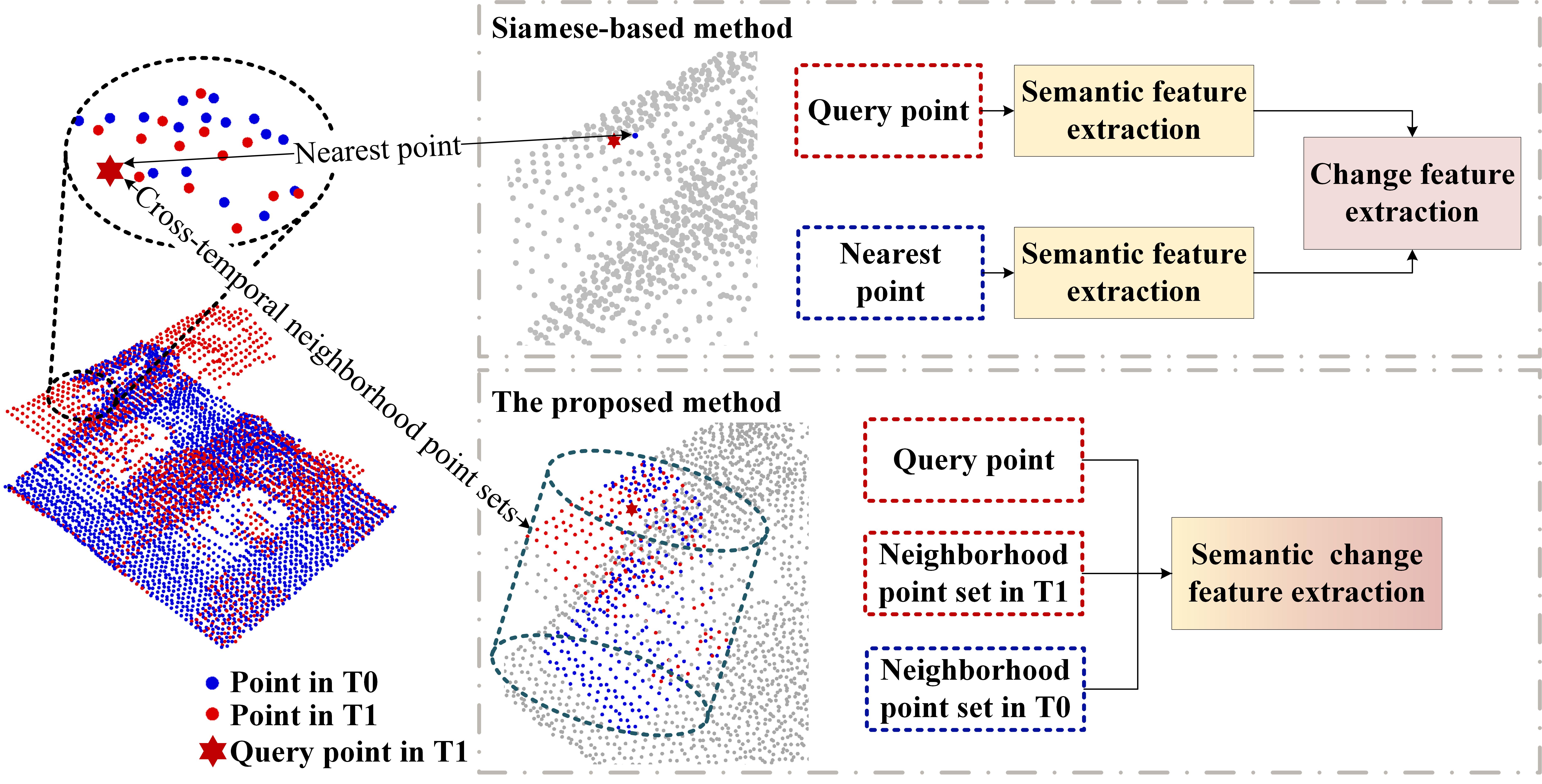}
    \caption{The difference between the proposed method and Siamese-based change detection networks in terms of neighborhood correspondence and feature extraction.}
    \label{fig_nei}
\end{figure*}

\section{Multi-Task Enhanced Cross-Temporal Point Transformer} \label{sec_method}


The workflow of the proposed ME-CPT network is illustrated in Fig. \ref{fig_workflow}. 
ME-CPT first performs cross-temporal serialization on the point clouds from two epochs, and dividing them into cross-temporal patches. 
Next, temporal indicators are introduced in the point embedding layer, and the embedded features are then fed into a multi-stage encoder-decoder, where each stage includes grid pooling and multi-depth attention block (Section. \ref{sec:ctpt}). The multi-task strategy is utilized to enhance feature discriminability (Section. \ref{sec:me}). 
The final output of ME-CPT includes point-level change prediction as well as semantic prediction for both epochs. 
The preliminaries of the basic point transformer are first introduced in Section \ref{sec:preliminary}.

\subsection{Preliminary of Point Transformer} \label{sec:preliminary}

Unlike 2D image change detection, 3D point cloud change detection faces challenges due to irregularity and intensive volume, making it hard to build a feature extraction network that captures long-range contextual relationships.
The Point Transformer (PT-V1) \citep{zhao2021point} processes point clouds to capture local and global geometric features, using self-attention mechanism to model spatial relationships and handle irregularity and scale variations.
To further expand the receptive field and improve computational efficiency, Point Transformer V3 (PT-V3) \citep{wu2024point} constructs a patch attention process by serializing point clouds. 

The self-attention mechanism(Section \ref{sec:pre_sam}) and point serialization (Section \ref{sec:pre_ser}) are the foundations of our method and are briefly reviewed below.


\subsubsection{Self-Attention Mechanism} \label{sec:pre_sam}


To adapt the transformer-based network \citep{vaswani2017attention} for processing point clouds, PT-V1 applies the self-attention mechanism to individual points, allowing the model to learn spatial relationships within the point cloud.
The self-attention mechanism calculates the attention weights for each point based on the relationships between its neighboring points, allowing the network to capture the spatial relationships and features. The point-based scaled dot-product attention is defined as shown in Eq. \eqref{eq1}:

\begin{equation}
y_i^{out} = \sum\limits_{{f_j} \in {\cal M}\left( i \right)} {\mathrm{Softmax} \left( {{{\varphi {{\left( {{f_i}} \right)}^ \top }\psi \left( {{f_j}} \right)} \mathord{\left/
 {\vphantom {{\varphi {{\left( {{f_i}} \right)}^ \top }\psi \left( {{f_j}} \right)} {\sqrt {{c_h}} }}} \right.
 \kern-\nulldelimiterspace} {\sqrt {{c_h}} }}} \right)\alpha } \left( {{f_j}} \right),\label{eq1}
\end{equation}
where $y_i^{out}$ is the output feature of the ${i}$-th point.  The subset ${{f_j} \in {\cal M}\left( i \right)}$ is a set of points features in a local neighborhood of the ${i}$-th point.  $\varphi$, $\psi$, and  $\alpha$ represent pointwize feature transformations, such as linear projections.  
$\frac{1}{{\sqrt {{c_h}} }}$ is the scaling factor.

\subsubsection{Point Serialization} \label{sec:pre_ser}
Due to the irregularity, discreteness, and large scale of point clouds, along with significant variations in point cloud density and object sizes, selecting appropriate neighborhoods is a significant challenge. 
Common neighborhood selection methods include k-nearest neighbors and fixed-radius spherical neighborhoods, but they struggle to adapt to variations in point cloud density.
To efficiently structure irregular point clouds, PT-V3 introduces point cloud serialization and a serialized attention mechanism, expanding each point's receptive field to 1024 points.

PT-V3 utilizes space-filling curves \citep{peano1990courbe}, such as Z-order curves \citep{morton1966computer} and Hilbert curves \citep{hilbertdritter}, to serialize points in specific spatial distributions.
Mathematically, a bijective function $\varphi: \mathbb{Z} \mapsto \mathbb{Z}{^n}$ is defined, where ${n}$ is the dimensionality of the space. Then, using a serialized encoding method, the points' coordinates are converted into integers to reflect their order along the space-filling curve. Using the inverse mapping function ${\varphi ^{ - 1}}:\mathbb{Z}{^n} \mapsto \mathbb{Z}$, the point cloud is serialized. PT-V3 assigns a 64-bit integer to each point to record the serialized code, allocating the trailing ${k}$ bits to the position encoded by ${\varphi ^{ - 1}}$ and the leading bits to the batch index ${b}$.
The formula for the serialization encoding is given by Eq. \eqref{eq2}:
\begin{equation}
\mathrm{Encode}\left( {p,b,g} \right) = \left( {b \ll k} \right)\left| {{\varphi ^{ - 1}}} \right.\left( {\left\lfloor {{p \mathord{\left/
 {\vphantom {p g}} \right.
 \kern-\nulldelimiterspace} g}} \right\rfloor } \right), \label{eq2}
\end{equation}
where  ${p}$  represents the point cloud coordinates,  ${g}$ represents the grid size, and ${\left\lfloor {{p \mathord{\left/
 {\vphantom {p g}} \right.
 \kern-\nulldelimiterspace} g}} \right\rfloor }$ represents the coordinates of the points converted to the grid.

\subsection{Cross-Temporal Point Transformer}\label{sec:ctpt}
The transformer-based structure and point cloud serialization offer an efficient approach for point cloud-based feature extraction. However, the challenge of urban multi-class semantic change detection lies in establishing comprehensive correspondence between multi-temporal points and extracting features containing semantic and change information.

\begin{figure*}[h]
    \centering
    \includegraphics[width=0.7\textwidth]{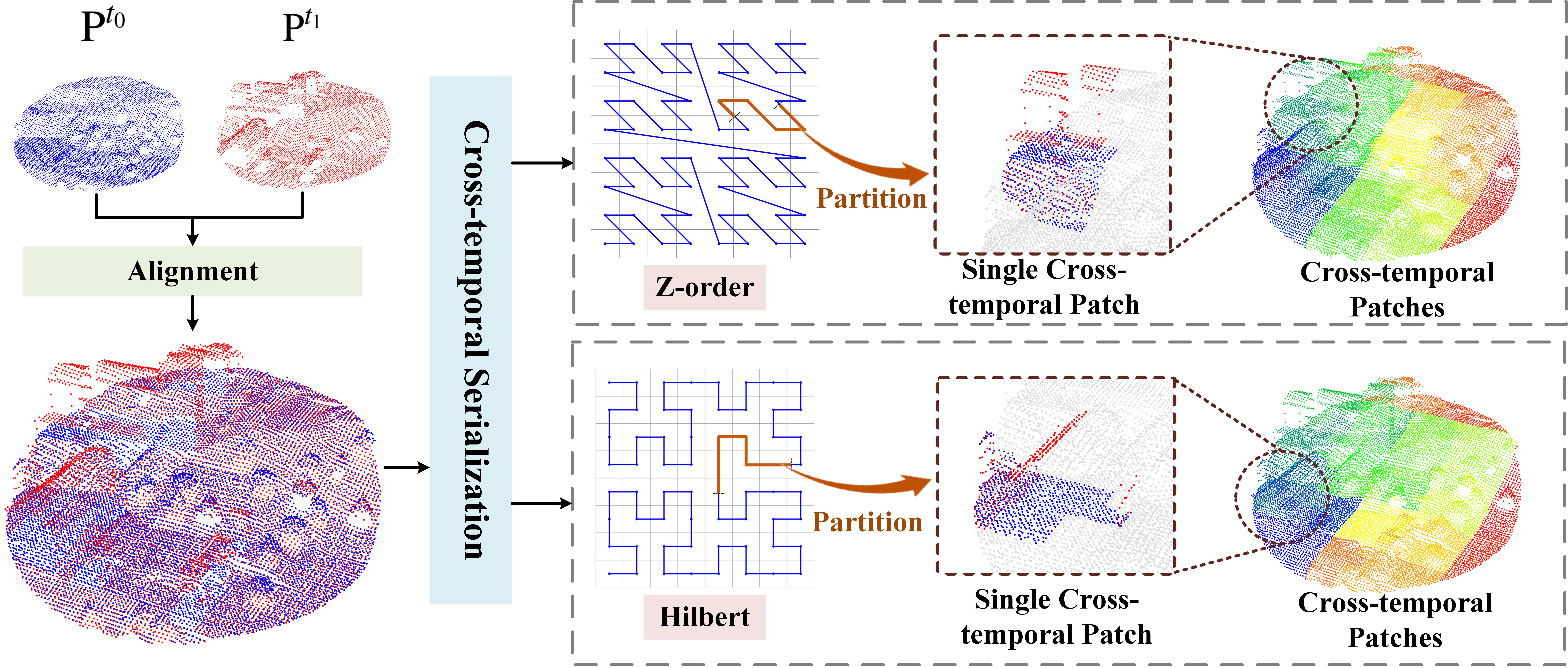}
    \caption{The process of point cloud serialization for cross-temporal patches using different space-filling curves.}
    \label{fig_serial}
\end{figure*}

\textbf{Remark 1.} \textit{Most existing works \citep{de_gelis_siamese_2023,zhan_pgn3dcd_2024,wang_end--end_2023} adopt Siamese structures, where two branches with shared weights independently compute semantic features for point clouds in each epoch. Change feature extraction in most of these works is achieved by matching points to their nearest or k-nearest neighbors between different epochs and then calculating change features by subtraction or concatenation. 
However, the extraction of change features remains limited because the complexity of multi-temporal distributional differences and semantic changes have not been fully considered.
Therefore, the lack of comprehensive multi-temporal spatial correspondences and the representation of change features will lead to difficulties in effectively extracting the changes that have occurred.} 

To improve change feature extraction, ME-CPT directly establishes spatiotemporal correspondences between multi-temporal point clouds and jointly extracts change features and semantic features, rather than adopting a Siamese structure.
Firstly, ME-CPT leverages cross-temporal serialization to construct the spatial distribution relationships of multi-temporal points and divides them into cross-temporal patches (Section. \ref{sec:ctps}). Then, for each patch, the cross-temporal attention mechanism  (Section. \ref{sec:ctpa}) is applied to extract semantic and change features. Finally, the encoder and decoder in the network framework are implemented by multi-stage, varying-depth cross-temporal attention block. The proposed method enhances the spatial correspondence and change feature extraction of multi-temporal point clouds. The differences between ME-CPT and the Siamese-based method are illustrated in 
Fig.\ref{fig_nei}.

\subsubsection{Cross-Temporal Point Serialization} \label{sec:ctps}

In this paper, for the change detection task, we innovatively align point clouds of two epochs and perform cross-temporal serialization within the same coordinate frame, along with implementing cross-temporal neighborhood partitioning, as shown in Fig.\ref{fig_serial}.
${\rm P}^{t_0}$ and ${\rm P}^{t_1}$ are point clouds of two epochs, aligned within the same coordinate frame:

\begin{subequations}
\begin{align}
    {{\rm P}^{{t_0}}} = \left\{ {{p^{{t_0}}}_n\left| {{p^{{t_0}}}_n \in \mathbb{R}^3,n = 1,2, \ldots ,N} \right.} \right\},\\
    {{\rm P}^{{t_1}}} = \left\{ {{p^{{t_1}}}_m\left| {{p^{{t_1}}}_m \in \mathbb{R}^3,m = 1,2, \ldots ,M} \right.} \right\}.
\end{align}
\end{subequations}

Then, the aligned point clouds are serialized with the serialized positional encoding shown in Eq.\eqref{eq2}, where $p \in {\rm P}^{t_0} \cup {\rm P}^{t_1}$.
Cross-temporal point clouds serialization uses space-filling curves to link the point clouds, constructing the spatiotemporal neighborhood relationships between all points. Specifically, cross-temporal serialization captures the spatial distribution of each point's neighborhood, including points both within and between different epochs.

To preserve sufficient spatial geometric relationships while minimizing computational resource requirements, the point cloud can be divided into multiple patches. By focusing on each patch individually, the network can capture the unique geometric and semantic features of the local space. 
Accordingly, ME-CPT adopts a cross-temporal patch partition approach, and the attention mechanism is implemented within each independent cross-temporal patch. This allows the model to capture both spatial and temporal dependencies within each cross-temporal patch, improving the ability to extract semantic and change features across different epochs.

As shown in Fig.\ref{fig_serial}, the cross-temporal serialization process uses different space-filling curve encoding to establish spatial relationships between point clouds from different epochs, linking cross-temporal point clouds based on their spatial relationships. 
Then, given the number of points ${k}$ in each patch, the serialized spatial connections are partitioned into non-overlapping cross-temporal patches.
Ultimately, each independent cross-temporal patch contains points from different epochs, where the different colors represent the independent cross-temporal patches. 


\begin{figure}[h]
    \centering
    \includegraphics[width=0.4\textwidth]{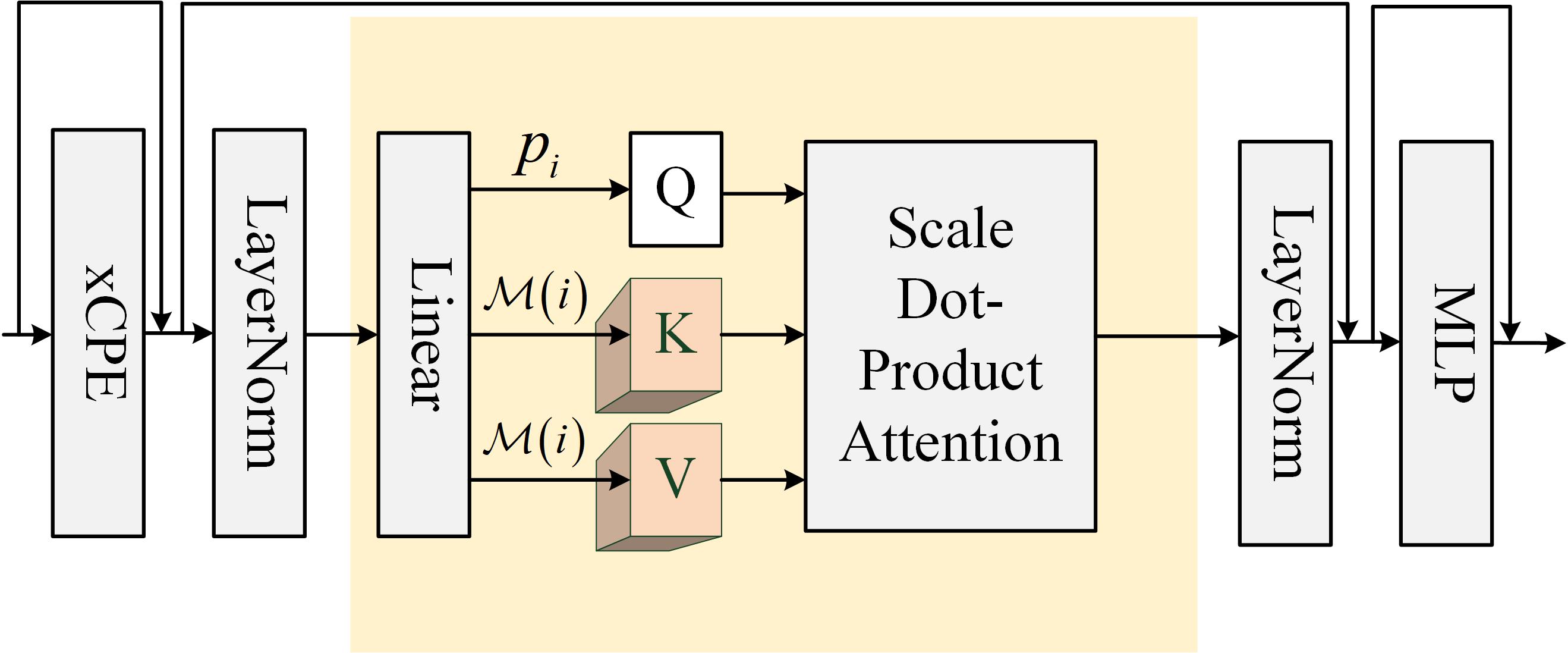}
    \caption{Cross-temporal Attention Block.}
    \label{fig_attenblock}
\end{figure}

\begin{figure*}[h]
    \centering
    \includegraphics[width=0.7\textwidth]{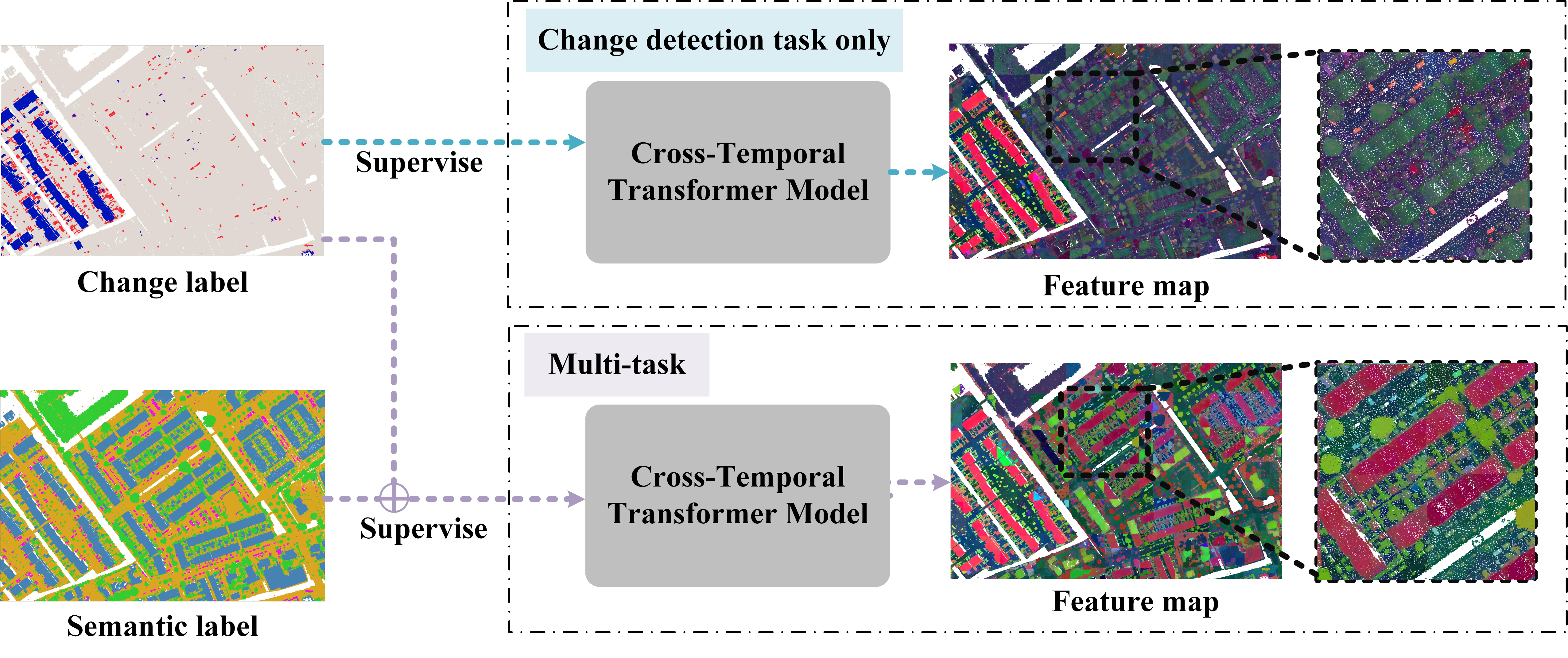}
    \caption{Multi-task enhanced feature extraction module.}
    \label{fig:multi-task}
\end{figure*}

\subsubsection{Cross-Temporal Patch Attention}  \label{sec:ctpa}

To distinguish points from different epochs within the same cross-temporal patch and enable the network to learn both semantic and change features for both the same and different epochs, ME-CPT introduces a temporal indicator as an additional input dimension. As shown in Fig.\ref{fig_workflow}, in the point embedding layer, in addition to the point coordinates ${x}$, ${y}$, ${z}$, a specific value is added as a temporal indicator for the points corresponding to different epochs. The temporal indicators allow the network to distinguish points from the two epochs and to learn both the temporal and spatial relationships effectively. The input of feature embedding layer is written as:

\begin{equation}
\left[ {\begin{array}{*{20}{l}}
{{\rm{ }}x_0^{{\rm{ }}{t_0}},{\rm{ }}y_0^{{\rm{ }}{t_0}},{\rm{ }}z_0^{{\rm{ }}{t_0}},...,{\rm{TI}_0}}\\
{....}\\
{{\rm{ }}x_N^{{\rm{ }}{t_0}},{\rm{ }}y_N^{{\rm{ }}{t_0}},{\rm{ }}z_N^{{\rm{ }}{t_0}},...,{\rm{TI}_0}}\\
{{\rm{ }}x_0^{{\rm{ }}{t_1}},{\rm{ }}y_0^{{\rm{ }}{t_1}},{\rm{ }}z_0^{{\rm{ }}{t_1}},...,{\rm{TI}_1}}\\
{....}\\
{{\rm{ }}x_M^{{\rm{ }}{t_1}},{\rm{ }}y_M^{{\rm{ }}{t_1}},{\rm{ }}z_M^{{\rm{ }}{t_1}},...,{\rm{TI}_1}}
\end{array}} \right], \label{eq3}
\end{equation}
where $\mathop x\nolimits_{}^{\mathop t\nolimits_0 },\mathop y\nolimits_{}^{\mathop t\nolimits_0 },\mathop z\nolimits_{}^{\mathop t\nolimits_0 } $ represent the coordinates of the points in the first epoch, and 
$\mathop x\nolimits_{}^{\mathop t\nolimits_1 },\mathop y\nolimits_{}^{\mathop t\nolimits_1 },\mathop z\nolimits_{}^{\mathop t\nolimits_1 } $ represent the coordinates of the points in the second epoch. 
${\rm{TI}_0}$ and ${\rm{TI}_1}$ denote the temporal indicators of the points respectively. 
In this paper, the temporal indicator value is defined as: $\left\{ {{\rm{TI}_t} = t,t \in  \{0,1\}} \right\}$.

Based on the reorganization of point clouds from two epochs into independent cross-temporal patches with spatiotemporal relationships, ME-CPT jointly extracts the change and semantic features of points through a cross-temporal attention mechanism. Accordingly, the encoder and decoder proposed in this method consist of multi-stage, varying-depth cross-temporal attention block, with each block containing a cross-temporal patch attention layer. As shown in Fig. \ref{fig_attenblock}, the cross-temporal attention block used in this paper is illustrated.

In cross-temporal attention layer, the attention mechanism is  applied to each independent cross-temporal patch. The attention mechanism within each patch is implemented using the scaled dot-product attention defined in Eq.\eqref{eq1}.  ${Q}$, ${K}$ and ${V}$ refer to the query, key, and value feature vectors obtained by applying linear projections to the point features. ${p_i}$ represents a point from any specific epoch, and ${\cal M}\left( i \right)$ is the set of all points from different epochs within the corresponding cross-temporal patch. Furthermore, before the attention layer, an enhanced conditional position encoding (xCPE) is introduced, which is implemented through a sparse convolution layer with skip connections. 
Additionally,in both the encoder and the decoder, before each stage of the varying-depth cross-temporal attention blocks, Grid Pooling and Grid UpPooling proposed by \citet{wu2022point} are integrated to perform downsampling and upsampling of the point cloud.

\subsection{Multi-task Enhanced Feature Extraction}\label{sec:me}

In the 3D semantic change detection, the sample distribution in the training data exhibits a significant imbalance, making it difficult for the network to learn sufficient discriminative features. To overcome the samples imbalance and effectively extract features, the proposed ME-CPT introduces a multi-task learning strategy, using semantic segmentation predictions for each epoch as an auxiliary task. 

More specifically, the proposed multi-task training strategy enhances the discriminability of features by separately feeding the features obtained from the decoder into the change detection branch and the semantic prediction branch. 
The change detection branch predicts the change categories for points in ${T_1}$ epoch, while the semantic segmentation branch predicts the semantic categories for points in both ${T_0}$ and ${T_1}$ epochs. 

In change detection tasks, the unchanged category dominates the majority of the samples and contains multiple types of land cover, while the spatial distribution of urban scenes and semantic changes are complex. By sharing the encoder and decoder and jointly training on different tasks, the model can learn more generalizable semantic features of different land covers through the auxiliary semantic segmentation task, thereby improving the understanding of semantic change information.
As shown in Fig. \ref{fig:multi-task}, adding the auxiliary semantic segmentation task significantly improves the model's ability to distinguish features of different land covers in unchanged regions.
The multi-task mechanism enhances the ability to extract multi-class land cover features, thereby mitigating the negative impact of sample imbalance on the representation of the unchanged regions.

\subsection{Training Loss}

To train the proposed ME-CPT network, the loss functions of the change detection branch and the semantic segmentation branch are jointly used, including the change detection loss ${{\cal L}_{cd}}$ and the semantic segmentation loss ${{\cal L}_{ss}}$. The overall loss function ${\cal L}_{total}$ composing semantic segmentation loss and the change detection loss is written as:
\begin{equation}
{{\cal L}_{total}} = \alpha {{\cal L}_{cd}} + \beta {{\cal L}_{ss}}, \label{eq:all}
\end{equation}
where $\alpha$ and $\beta$ are the weights for the semantic segmentation branch and the change detection branch, respectively. 
In the experiment, both ${\alpha}$ and ${\beta}$ are set to ${0.5}$. More specifically, the change detection loss ${{\cal L}_{cd}}$ and the semantic segmentation loss ${{\cal L}_{ss}}$ are calculated using Eq.\eqref{eq:loss}.
\begin{equation}
{\cal L} =  - \frac{1}{C}\sum\limits_{i = 1}^C {{l_i}\log \left( {{pred_i}} \right)} , \label{eq:loss}
\end{equation}
where $C$ is the number of semantic classes or change detection classes. ${{l_i}}$ and ${{pred_i}}$ represent the ground truth and the predicted labels, respectively.

\section{New York City Semantic Change Detection Dataset (NYC-SCD)}  \label{sec_dataset}
To the best of our knowledge, currently there is no publicly available urban 3D change detection dataset based on ALS data. To address this gap, we propose a more challenging multi-class 3D semantic change detection dataset.
\subsection{Data acquisition}\label{sec:dataac}
\begin{figure}[h]
    \centering
    \includegraphics[width=0.45\textwidth]{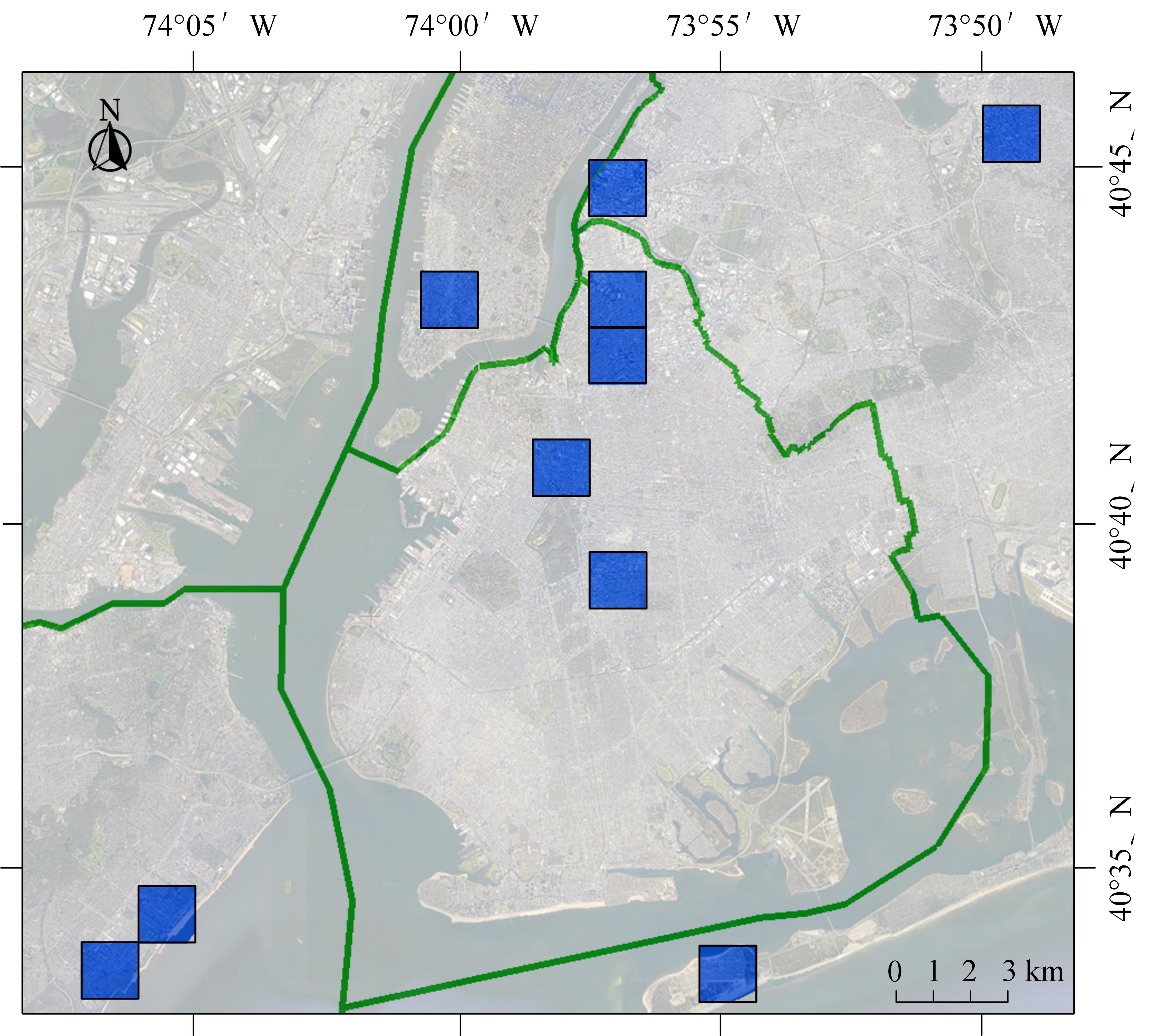}
    \caption{The location of the selected NYC data.}
    \label{fig_nycgeo}
\end{figure}

The new republic 3D semantic change detection dataset is sourced from high-quality ALS data collected in 2014 and 2017, released by the New York City agency. The two epochs of ALS point cloud data cover a wide range of scenes, and the variety of change samples is sufficient to support research in 3D change detection. 
The 2014 ALS data were acquired using the Leica ALS70, achieving a root mean square error (RMSE) accuracy of 5.3 $cm$ and a point density of 5.9 $points/m^2$ based on ground control measurements. The 2017 ALS data were captured using the Leica ALS80, with an improved RMSE of 3.5 $cm$ and a point density of 8 $points/m^2$. Taking into account the diversity of the change samples and urban scenes, we selected 10 scenes, each covering an area of $1.5 \times 1.5 km^2$, distributed across the five boroughs of New York City. The total area covered by these scenes is 22.5 $km^2$, as shown in Fig. \ref{fig_nycgeo}, where the blue areas represent the selected data coverage locations.

\subsection{Data annotation}
This paper annotates the semantic change detection dataset for the selected multi-temporal point cloud data described in Section \ref{sec:dataac}. In NYC-SCD, the ALS point clouds from both epochs are annotated with semantic labels, and the ${T_1}$ epoch point clouds are additionally annotated with semantic change labels.

The semantic labels of the two epoch point clouds are annotated into four categories, with the following classification for each category:
\begin{itemize}
  \item Ground: road surfaces, pavement, flat terrain, etc.
  \item Building: all man-made structures.
  \item Vegetation: trees and other low-growing plants.
  \item Clutter: ground objects that are not vegetation.
\end{itemize}

Based on the semantic labels and rules for change detection, the ${T_1}$ epoch point cloud is annotated with four change categories. The categories are as follows:
\begin{itemize}
  \item Unchanged: all semantic categories that have not undergone any changes.
  \item Newly built: newly constructed buildings and added roof attachments.
  \item Demolition: areas where buildings have been demolished.
  \item New clutter: newly appearing objects that include semantic categories such as vegetation and clutter.
\end{itemize}

Additionally, points that represent rare categories in the dataset, which are not relevant to the urban semantic change categories, are manually checked and labeled as "Ignored" categories. These include features such as railways, shoreline shrubs, and surface water points.

\subsection{Statistics of NYC-SCD}
\begin{figure}[h]
    \centering
    \includegraphics[width=0.5\textwidth]{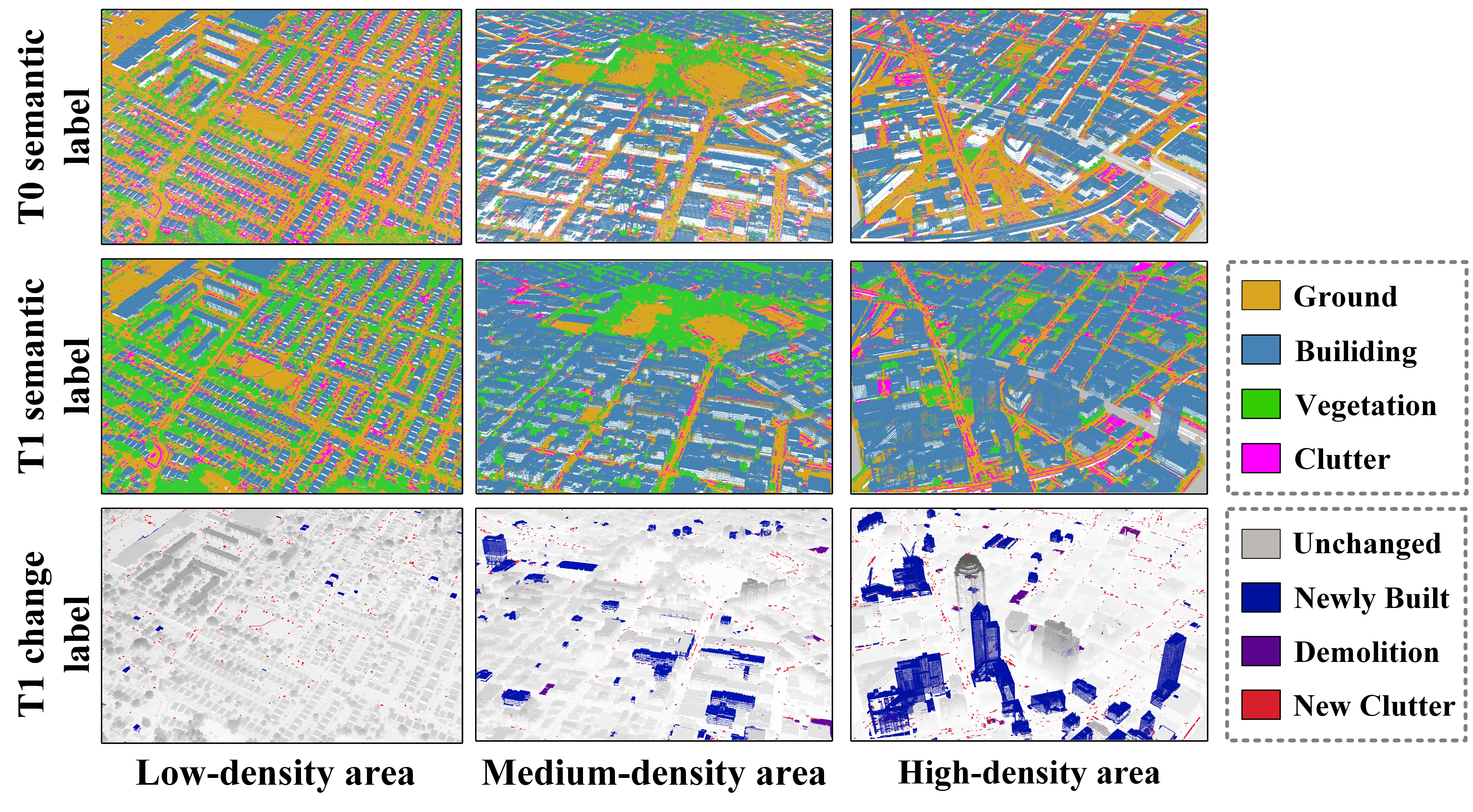}
    \caption{Illustration of multi-temporal semantic labels and change labels for different scene types.}
    \label{fig_nycclass}
\end{figure}

\begin{figure}[h]
    \centering
    \includegraphics[width=0.35\textwidth]{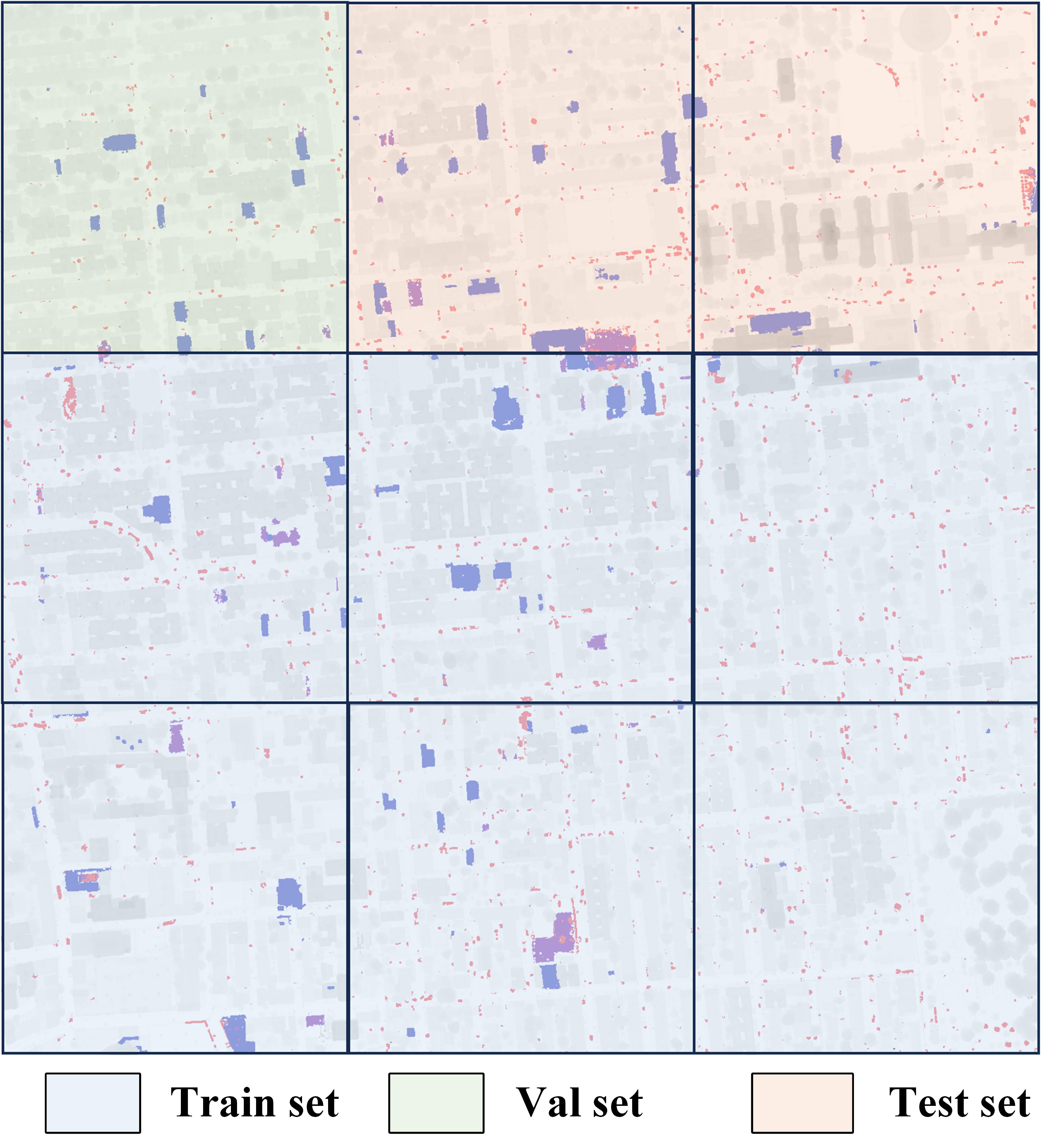}
    \caption{Illustration of the training, validation, and test set division for a single scene.}
    \label{fig_nycsplit}
\end{figure}

The 10 selected scenes exhibit significant diversity, representing typical urban forms from various cities. Based on the distribution of building heights and density, these scenes can be broadly categorized into three types: high-density, medium-density, and low-density areas. High-density areas feature numerous high-rise buildings with dense distributions. Medium-density areas consist of buildings with moderate heights (within 20 floors) and relatively even distribution. Low-density areas are residential zones located away from commercial or office areas, characterized by lower buildings and abundant open spaces and green areas. 
Examples illustrating the semantic and change labels for three different types of scenes are presented in Fig.\ref{fig_nycclass}. To account for diversity between scenes, each scene was divided into train, validation, and test sets in a 6:1:2 ratio, as illustrated in Fig.\ref{fig_nycsplit}. This division ensures a balanced distribution, providing an appropriate proportion for model training, validation, and evaluation.

The point counts of semantic labels and change labels  for the NYC-CD dataset are summarized in Tab.\ref{tab:datasta}.

\begin{table*}[]
\caption{Experimental Parameter Settings of the Proposed Method on Different Datasets. \label{tab:datasta}}
\centering
\scalebox{0.9}{
\begin{tabular}{cccccccccc}
\hline
\multirow{2}{*}{Dataset} & \multirow{2}{*}{Subset} & \multicolumn{4}{c}{Points for different categories of semantic} & \multicolumn{4}{c}{Points for different categories of change} \\ \cline{3-10} 
                         &                         & Ground         & Building      & Vegetation     & Clutter       & Unchanged         & Newly Built       & Demolition        & New Clutter       \\ \hline
\multirow{3}{*}{${T_0}$ Epoch} & Train                  & $1.31\times10^8$ & $1.25\times10^8$ & $5.19\times10^7$ & $1.02\times10^7$ & $3.07\times10^8$   & $7.41\times10^6$   & $1.99\times10^6$  & $1.54\times10^6$  \\ 
                         & Val                    & $2.21\times10^7$ & $1.98\times10^7$ & $1.04\times10^7$ & $1.80\times10^6$ & $5.22\times10^7$   & $1.32\times10^6$   & $2.91\times10^5$  & $3.17\times10^5$  \\ 
                         & Test                   & $4.69\times10^7$ & $4.64\times10^7$ & $2.52\times10^7$ & $3.42\times10^6$ & $1.18\times10^8$   & $2.25\times10^6$   & $6.49\times10^5$  & $5.78\times10^5$  \\ \hline
\multirow{3}{*}{${T_1}$ Epoch} & Train                  & $3.28\times10^7$ & $2.41\times10^7$ & $4.97\times10^6$ & $4.21\times10^6$ & \multicolumn{4}{c}{-}                                   \\ 
                         & Val                    & $5.32\times10^6$ & $3.75\times10^6$ & $8.14\times10^5$ & $6.56\times10^5$ & \multicolumn{4}{c}{-}                                   \\ 
                         & Test                   & $1.18\times10^7$ & $8.33\times10^6$ & $1.94\times10^6$ & $1.52\times10^6$ & \multicolumn{4}{c}{-}                                   \\ \hline
\end{tabular}
}
\end{table*}

\section{Experiment} \label{sec_experiment}

\subsection{Implementation details}
\textbf{Model architecture:}
The network proposed in this paper adopts a four-stage encoder-decoder architecture, with the block depths 
$\left[ {2,2,6,2} \right]$ and $\left[ {2,2,2,2} \right]$ respectively. The patch size is set to 1024 for  datasets URB3DCD-V2, AHN-CD, and NYC-SCD, while for the street-level dataset SLPCCD, it is set to 512.

\textbf{Traning parameters:} 
In the training process, the point clouds of both epochs are first voxelized using a voxel size ${v}$. Then, the cylinder-based sample sampling method is used, where each input sample to the network consists of points from a cylindrical region with a given radius ${r}$. The training parameters and optimization settings for each dataset are listed in Tab.\ref{tab:parameters}.

\begin{table*}[]
\caption{Experimental Parameter Settings of the Proposed Method on Different Datasets. \label{tab:parameters}}
\centering
\scalebox{0.9}{
\begin{tabular}{lllllllll}
\hline
  Dataset    & Voxel size & Cylinder Radius & Train samples & Val samples & Train Epochs & Optimizer & LR Schedule \\ \hline
URB3DCD-V2 & 0.3        & 50              & 6000          & 1000        & 100          & AdamW     & OneCircleLR \\
AHN-CD     & 0.5        & 25              & 6000          & 500         & 100          & AdamW     & OneCircleLR \\
NYC-SCD    & 0.5        & 25              & 6000          & all         & 100          & AdamW     & OneCircleLR \\
SLPCCD     & 0.03       & -               & all           & all         & 200          & SGD       & MultiSteoLR        
           
           \\ \hline
\end{tabular}}
\end{table*}

\subsection{Evaluation Metric}
This paper uses widely adopted evaluation metrics, including overall accuracy (OA), intersection over union (IoU), and mean IoU (mIoU), for point-wise accuracy assessment.

\subsection{3D Chaneg Detection Datasets}
To validate the effectiveness of the proposed method, experiments are conducted not only on the proposed NYC-SCD dataset, but also on two publicly available datasets with change annotations (Ur3DCD-V2 and SLPCCD), as well as a multi-temporal ALS point cloud dataset with semantic annotations (AHN-CD [Actueel Hoogtebestand Nederland Change Detectio]).The descriptions of the datasets are detailed as follows.

\subsubsection{Simulated urban change detection dataset (URB3DCD-V2)}

\citet{de2021change} employees a simulator based on the LoD2 model of Lyon, France, to generate multi-temporal airborne point cloud data by simulating the addition and removal of buildings and assigning change labels. Based on this dataset, \citet{de_gelis_siamese_2023} improves it by randomly incorporating vegetation, vehicles, and other moving objects, resulting in a multi-class change detection dataset. This dataset includes semantic labels for point clouds of two epochs  ${T_0}$ and ${T_1}$, as well as change labels for ${T_1}$. In the proposed paper, experiments are conducted on the first subset of the dataset, with a point density of approximately 0.5 $points/m^2$. The semantic labels of point clouds are ground, building, vegetation and mobile objects, while the change labels include unchanged, newly built, demolition, vegetation growth, new vegetation, missing vegetation, and mobile objects.

\subsubsection{Street-Level Point Clouds Change Detection Dataset (SLPCCD)}

SLPCCD is a street-level change detection dataset proposed by \citet{wang_end--end_2023}.
The data consists of colored point clouds collected using mobile LiDAR in 2016 and 2020. In the SLPCCD dataset, the data are divided into cylinders with a radius of 3 meters, centered on street-level objects, and labeled as either change points or background points. For the ${T_0}$ epoch, the change point class is labeled as add, while for the ${T_1}$ epoch, the class is labeled as removed. The objects involved in the changes in this dataset include streetlights, signposts, benches, pedestrians, and other street-level objects.

\subsubsection{Netherlands Change Detection (AHN-CD)}

The AHN-CD dataset is sourced from multiple national LiDAR acquisitions released by the Netherlands. This study uses data from the third and fourth acquisitions, AHN3 and AHN4. The data coverage area is consistent with \citep{de_gelis_siamese_2023}, with the train, validation, and test sets covering approximately 3.75 $km^2$, 1.25 $km^2$, and 2.5 $km^2$, respectively. The point density for AHN3 ranges between 10 and 14 points/m², while for AHN4, it ranges from 20 to 24 points/m². In this work, the semantic labels for AHN3 and AHN4 are relabeled as ground, building, vegetation, and clutter. Following the change labeling process proposed by \citet{de_gelis_siamese_2023}
the points in the  ${T_1}$  epoch were annotated as unchanged, newly built, demolition, and new clutter.

\subsection{Experimental Results}

\subsubsection{Simulated urban change detection dataset (Urb3DCD-V2)}

\begin{table*}[]
\caption{Quantitative evaluation results on the simulated dataset (Urb3DCD-V2). \label{tab:moni}}
\centering
\scalebox{0.85}{
\begin{tabular}{lllllllll}
\hline
                    & \multicolumn{1}{c}{\multirow{2}{*}{mIoU(\%)}} & \multicolumn{7}{c}{IoU(\%)}                                                                                                 \\ \cline{3-9} 
                    & \multicolumn{1}{c}{}                          & Unchanged      & New Building   & Demolition     & Vegetation Growth & Missing Vegetation & New Vegetation & Mobile Obiects \\ \hline
RF \citep{tran_integrated_2018}                  & 52.37                                         & 92.72          & 73.16          & 64.6           & 75.17             & 19.78              & 7.78           & 73.71          \\
Siamkpconv \citep{de_gelis_siamese_2023} & 80.12                               & 95.82 & 86.67 &78.66 & 93.16    & 65.18              & 65.46          & 91.55          \\
PGN3DCD  \citep{zhan_pgn3dcd_2024}           & 88.12                                         & 97.77          & \textbf{96.93}          & 83.08          & \textbf{96.82 }            & \textbf{73.00}               & 76.01          & 92.56          \\
\textbf{ours}       & \textbf{88.56}                                & \textbf{97.86} & 96.20 & \textbf{85.4}  & 95.39    & 71.76     & \textbf{78.44} & \textbf{94.90} \\ \hline
\end{tabular}}
\end{table*}

\begin{figure}[h]
    \centering
    \includegraphics[width=0.45\textwidth]{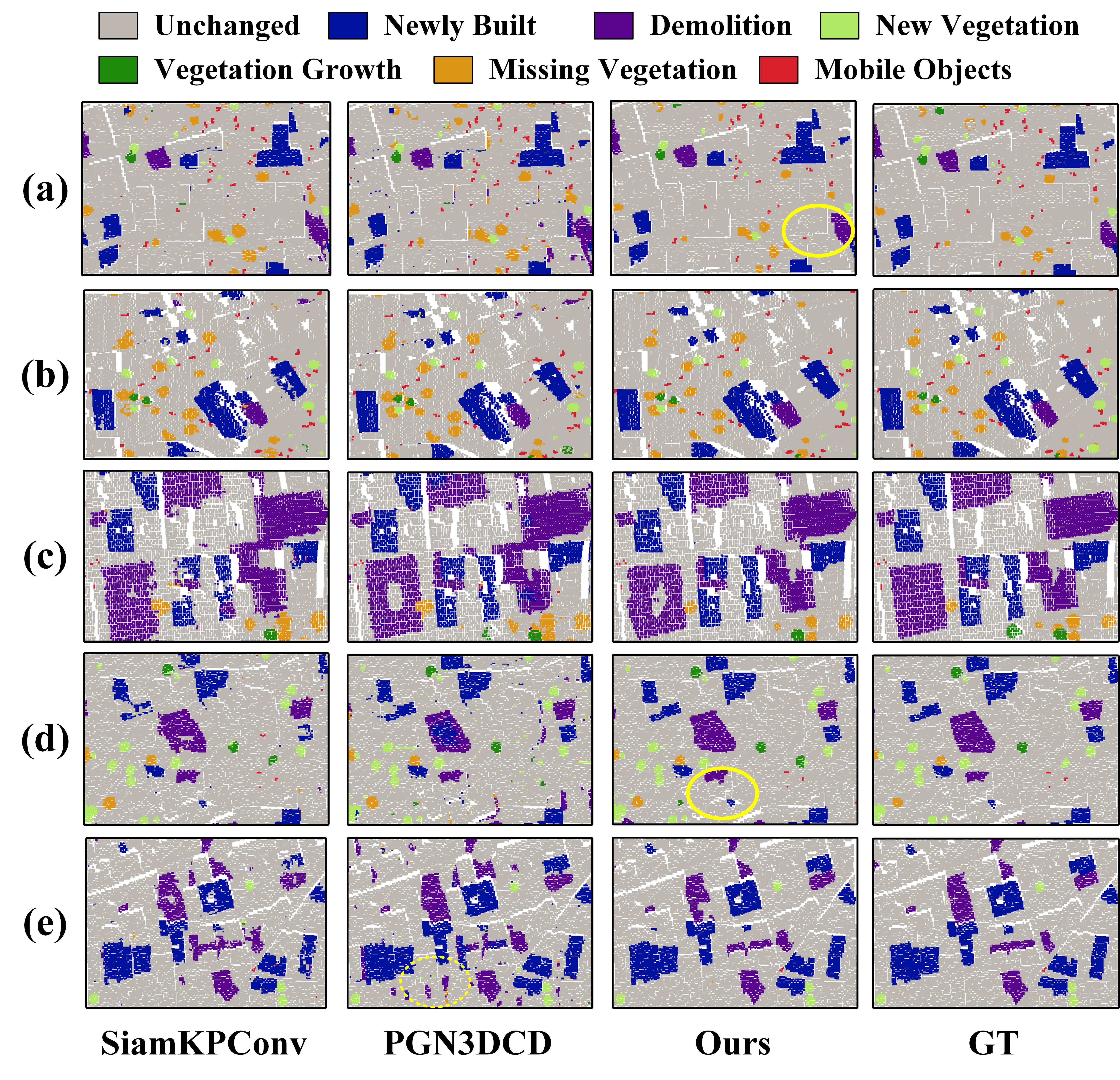}
    \caption{Qualitative results on the simulated dataset (Urb3DCD-V2).}
    \label{fig_moni}
\end{figure}

The quantitative evaluation and qualitative results of the proposed method, along with the comparison methods, are presented in Fig.\ref{fig_moni} and Tab.\ref{tab:moni}. The proposed method achieves the best mIoU in the simulated dataset. Among the comparison methods, RF \citep{tran_integrated_2018} relies on handcrafted features, while SiamKPConv \citep{de_gelis_siamese_2023} and PGN3DCD \citep{zhan_pgn3dcd_2024} are deep learning-based change detection approaches. 

The results demonstrate that deep learning-based methods have a significant advantage in multi-class change detection, as the diversity of land cover categories makes it challenging to achieve accurate detection using manually designed features. Furthermore, the regularity and high quality of the simulated dataset contribute to the relatively high accuracy of all deep learning methods.

Compared to other approaches, the proposed method benefits from its larger receptive field, allowing it to extract larger-scale features and significantly reduce false detections, as illustrated in Fig.\ref{fig_moni} (a) and (e). Moreover, as demonstrated in Fig.\ref{fig_moni} (d), the method achieves superior completeness in the extraction of change objects, with notably improved boundary preservation, enhancing overall change detection quality.

\subsubsection{Street-Level Point Clouds Change Detection Dataset (SLPCCD)}

\begin{figure}[h]
    \centering
    \includegraphics[width=\linewidth]{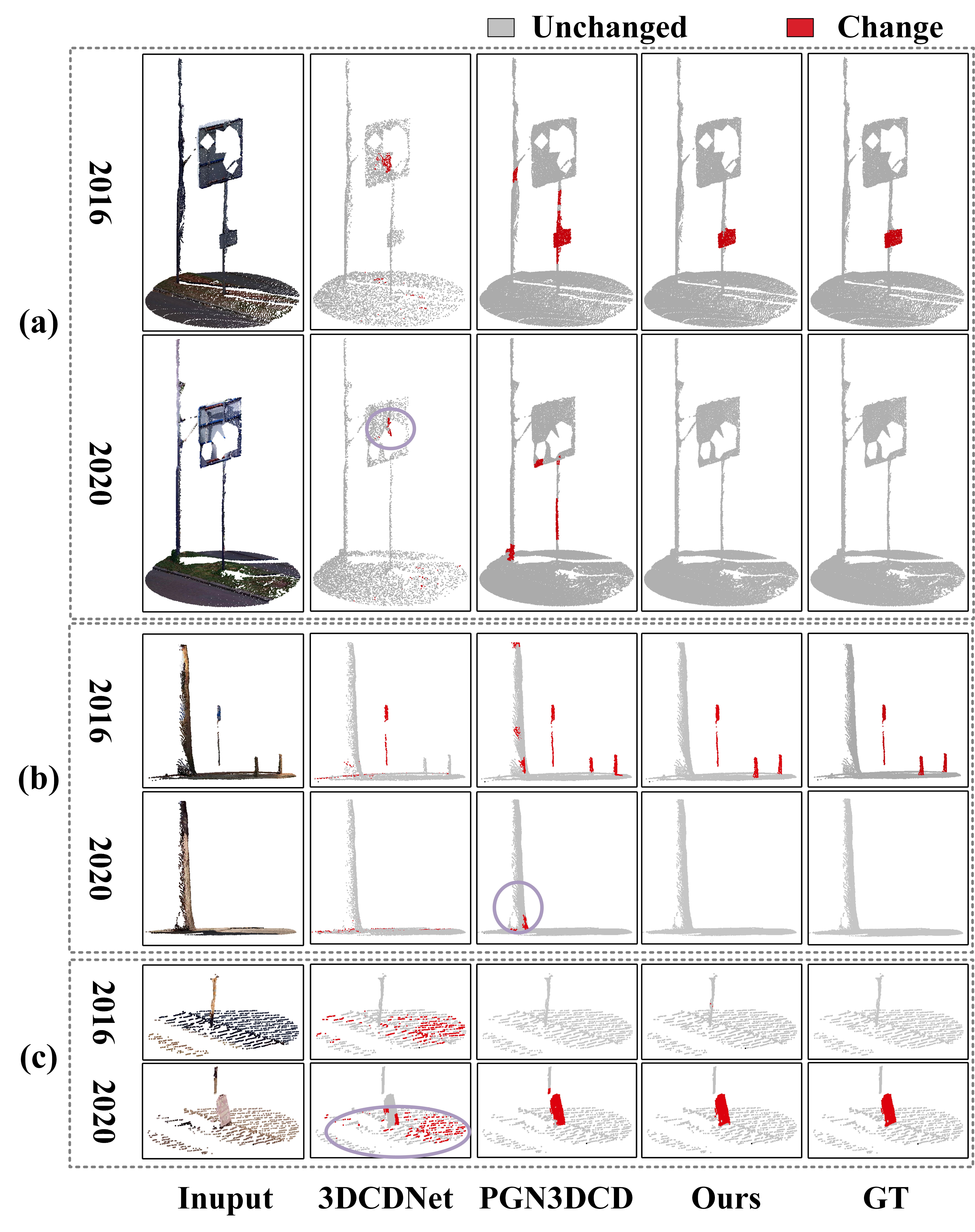}
    \caption{Qualitative results on the street-level point clouds change detection dataset (SLPCCD).}
    \label{fig_shrec}
\end{figure}

The quantitative evaluation and qualitative results on the street-level dataset, along with the comparison methods, are presented in Fig.\ref{fig_shrec} and Tab.\ref{tab:shrec}. The proposed method demonstrates the highest accuracy across all categories. Change detection in the street-level dataset primarily emphasizes recognizing small objects and changes in the local street scene, without differentiating specific types of change. However, due to variations in revisit scan angles and equipment, the task requires leveraging higher-level global features to accurately distinguish real changes from geometric errors caused by data inconsistencies. The proposed method effectively addresses these challenges and achieves superior performance.

As shown in Fig.\ref{fig_shrec}(a), when there are scan differences in the data, existing methods tend to misinterpret these discrepancies as changes, while the proposed method accurately predicts them as background points. Therefore, based on the street-level dataset experiment, it is clear that the differences in local features require geometric long-distance global features to enhance the accuracy of change detection.

\begin{table}[]
\caption{Quantitative results on the street-level point clouds change detection dataset (SLPCCD). \label{tab:shrec}}
\centering
\scalebox{0.9}{
\begin{tabular}{lllllllll}
\hline
\multicolumn{1}{c}{}                 & \multirow{2}{*}{OA(\%)} & \multirow{2}{*}{mIOU(\%)} & \multicolumn{3}{c}{IoU(\%)}                      \\ \cline{4-6} 
\multicolumn{1}{c}{}                 &                         &                           & Unchanged      & Removed        & Added          \\ \hline
DGCNN-based \citep{wang_end--end_2023} & 94.07                   & 62.3                      & 93.76          & 43.73          & 49.4           \\
3DCDNet \citep{wang_end--end_2023}     & 95.85                   & 74.45                     & 95.52          & 63.46          & 64.37          \\
PGN3DCD \citep{zhan_pgn3dcd_2024}      & 96.97                   & 82.17                     & 96.67          & 74.58          & 75.25          \\
\textbf{ours}                        & \textbf{98.1}           & \textbf{84.53}            & \textbf{97.96} & \textbf{79.08} & \textbf{76.55}
\\ \hline
\end{tabular}}
\end{table}

\subsubsection{Netherlands Change Detection (AHN-CD)}

\begin{figure*}[h]
    \centering
    \includegraphics[width=0.9\textwidth]{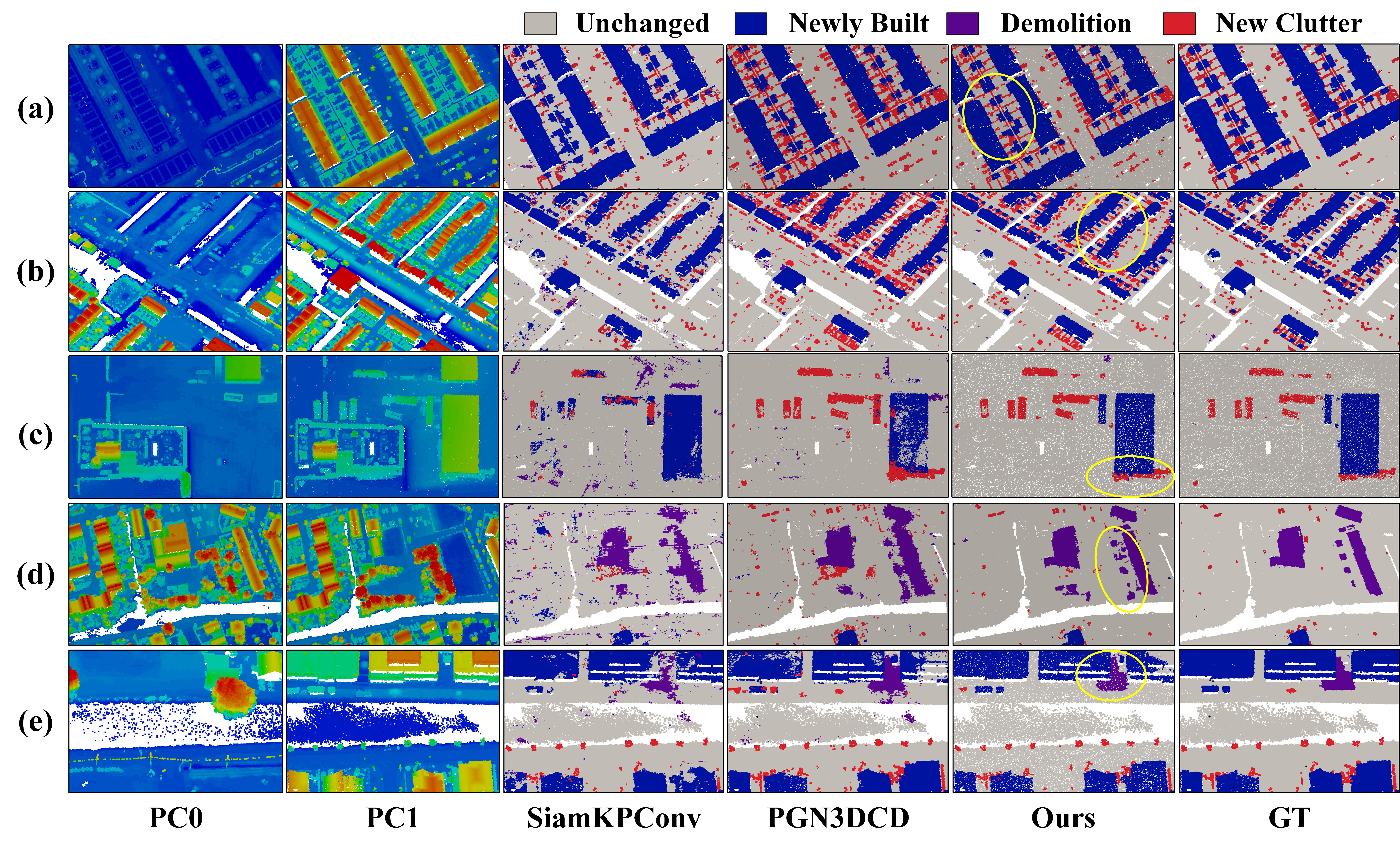}
    \caption{Qualitative results on the Netherlands Change Detection (AHN-CD).}
    \label{fig_ahn}
\end{figure*}

The proposed method was compared with comparison methods in the AHN-CD dataset, and the quantitative evaluation and qualitative results are shown in Tab.\ref{tab:ahn} and Fig.\ref{fig_ahn}. Due to the complexity of changes in urban area, existing methods struggle to consistently annotate all seven categories, and there is greater focus on building changes or the emergence of small objects.
From the quantitative evaluation, it is evident that the mean accuracy for change categories is significantly lower on the AHN-CD dataset compared to the simulated dataset. Variations in flight paths, scan angles, partial occlusions, and errors from equipment make it challenging to distinguish real changes from pseudo-changes in unchanged areas across different epochs. Consequently, focusing solely on local change features can result in more false detections.

\begin{table}[]
\caption{Quantitative results on the Netherlands Change Detection (AHN-CD).  \label{tab:ahn}}
\centering
\scalebox{0.75}{
\begin{tabular}{lcccccc}
\hline
 & \multirow{2}{*}{OA(\%)} & \multirow{2}{*}{mIoU(\%)} & \multicolumn{4}{c}{IoU(\%)}                                      \\ \cline{4-7} 
 &                         &                           & Unchanged      & Newly Built    & Demolition    & New Clutter    \\ \hline
3DCDNet \citep{wang_end--end_2023}       & 93.59                   & 45.66                     & 93.75          & 65.19          & 2.19          & 21.53          \\
SiamKPconv \citep{de_gelis_siamese_2023} & 94.25                   & 57.8                      & 94.18          & 72.53          & 36.72         & 27.78          \\
PGN3DCD \citep{zhan_pgn3dcd_2024}        & 95.82                   & 66.76                     & 95.8           & 83.88          & 45.74         & 41.63          \\
\textbf{ours}                            & \textbf{96.63}          & \textbf{75.94}            & \textbf{96.56} & \textbf{85.55} & \textbf{65.1} & \textbf{56.55}
\\ \hline
\end{tabular}}
\end{table}

In urban real-world scenarios, the shapes and sizes of buildings, vegetation, and ground objects vary greatly, making semantic change detection more challenging. For example, as shown in Fig.\ref{fig_ahn} (a) and (b) (highlighted with yellow circles), the proposed method accurately identifies new buildings or new clutter based on the $T_1$ epoch land-cover class, demonstrating better performance in distinguishing different types of changes. Similarly, as shown in Fig.\ref{fig_ahn} (c), when clutter like vegetation near new buildings, determining the boundary between new buildings and clutter becomes difficult. However, the proposed method effectively distinguishes between these categories.

Urban areas often exhibit complex changes, with simultaneous additions and removals of different objects in the same region. Among the change types in the AHN-CD dataset, apart from newly added buildings and clutter, more attention is paid to the removal of building-like objects. As shown in  Fig.\ref{fig_ahn} (d) and (e), when both vegetation and buildings are removed in the same region, the task requires accurate identification of the removed building area in the  $T_0$ epoch. Accurate detection of building removals relies on precise extraction of semantic features from the  $T_0$ epoch. The proposed multi-task mechanism significantly enhances the distinguishability of different objects, leading to more accurate recognition of demolished buildings.
\subsubsection{New York City Semantic Change Detection Dataset (NYC-SCD)}

\begin{figure*}[h]
    \centering
    \includegraphics[width=0.9\textwidth]{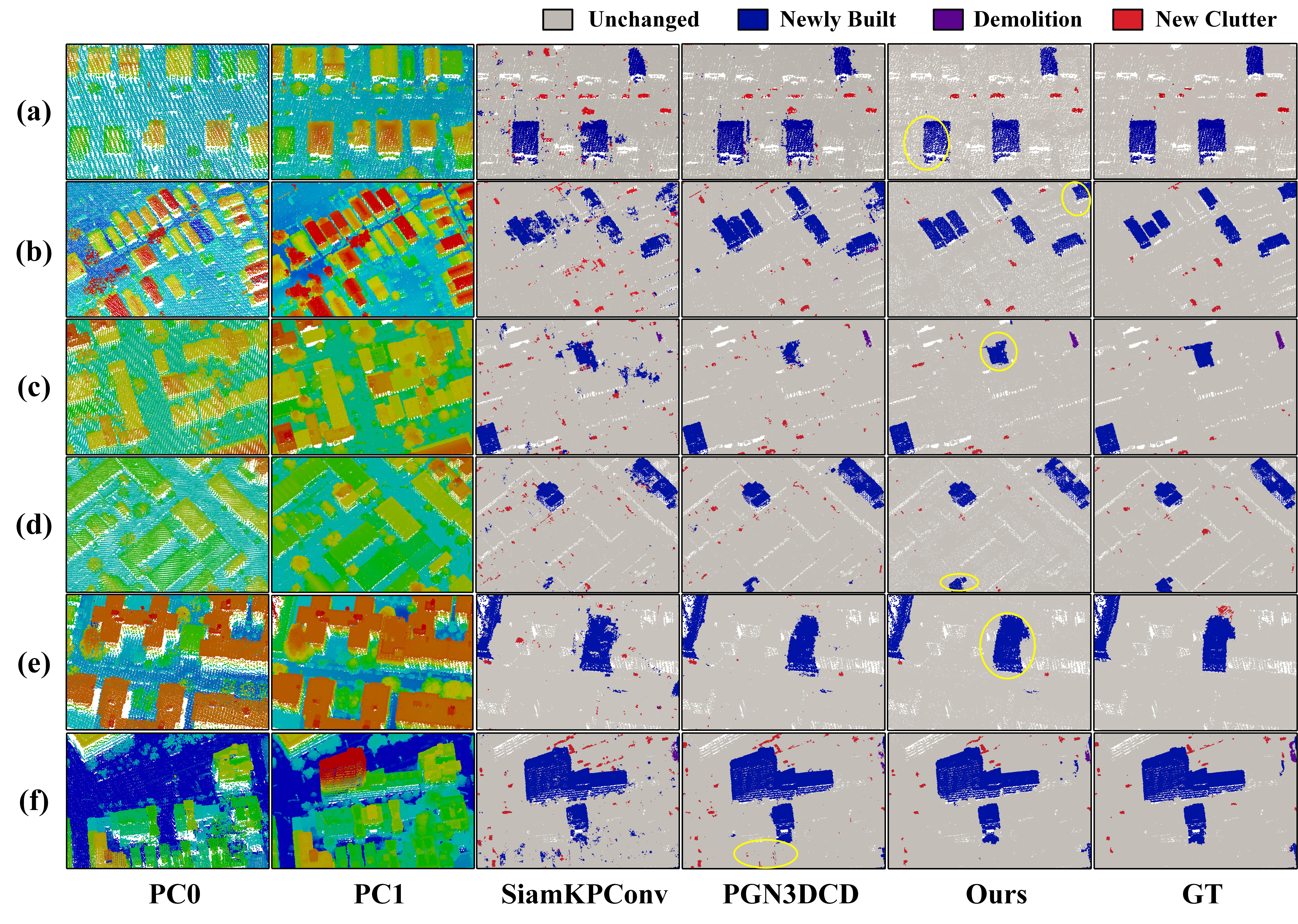}
    \caption{Qualitative results on the New York City Semantic Change Detection Dataset (NYC-SCD).}
    \label{fig_nyc}
\end{figure*}

Currently, no publicly available multi-temporal ALS point cloud dataset exists for multi-class 3D semantic change detection. To address this gap, this article presents a semantic change detection dataset spanning 22.5 $km^2$ of various urban scenes, annotated with semantic labels in two epochs. Using this dataset, both the proposed method and the existing approaches were evaluated, with quantitative results summarized in Tab.\ref{tab:nyc} and visualized in Fig.\ref{fig_nyc}. The proposed method demonstrated superior performance, achieving the highest accuracy in identifying change categories in the NYC-SCD dataset.

As shown in Fig.\ref{fig_nyc}, the proposed method performes effectively across various scenarios with different building densities. However, the accuracy for the "demolition" category is lower than that for the "newly built" category. This discrepancy arises because the "demolition" label is assigned to the remaining ground points in ${T_1}$, requiring the network to accurately infer the semantic context of the corresponding neighborhood points ${T_0}$ and to interpret the changes effectively.

\begin{table}[]
\caption{Quantitative results on the New York City Semantic Change Detection Dataset (NYC-SCD). \label{tab:nyc}}
\centering
\scalebox{0.75}{
\begin{tabular}{lllllll}
\hline      
 & \multicolumn{1}{c}{\multirow{2}{*}{OA(\%)}} & \multicolumn{1}{c}{\multirow{2}{*}{mIoU(\%)}} & \multicolumn{4}{c}{IoU(\%)}                              \\ \cline{4-7} 
 & \multicolumn{1}{c}{}                        & \multicolumn{1}{c}{}                          & \multicolumn{1}{c}{Unchanged} & \multicolumn{1}{c}{Newly Built} & \multicolumn{1}{c}{Demolition} & \multicolumn{1}{c}{New Clutter} \\ \hline
SiamKPconv \citep{de_gelis_siamese_2023} & 97.01                                       & 50.93                                         & 97.09                         & 47.72                           & 40.46                          & 8.45                            \\
PGN3DCD \citep{zhan_pgn3dcd_2024}        & 98.26                                       & 60.95                                         & \textbf{98.33}                & 71.99                           & 49.82                          & 24.47                           \\
\textbf{ours}                            & \textbf{98.78}                              & \textbf{66.17}                                & 97.65                         & \textbf{74.93}                  & \textbf{60.27}                 & \textbf{31.82}       
\\ \hline    \end{tabular}} 
\end{table}

The accuracy of the "new clutter" category is relatively low due to the diverse types of objects it encompasses, including vegetation, vehicles, and other non-building targets. Additionally, small-sized "new clutter" changes are particularly vulnerable to pseudo-changes caused by discrepancies between multi-temporal scans. As shown in Fig.\ref{fig_nyc} (c), the proposed method produces more complete change detection results while exhibiting fewer false positives.

Unlike the AHN-CD dataset, where most new buildings are standalone structures, high-density areas in the NYC-SCD dataset often feature new buildings tightly connected to existing structures due to limited urban space. Fig.\ref{fig_nyc} (e) and (f) illustrate these complex connections, where the proposed method achieves superior completeness in identifying such changes.

In high-density urban areas, characterized by taller buildings and narrower spacing, the impact of inter-temporal scan differences is more pronounced. As shown in Fig.\ref{fig_nyc} (f), in addition to localized geometric differences, capturing long-range global features is essential for distinguishing true changes from pseudo-changes. The proposed algorithm effectively integrates long-range features, enabling it to better differentiate genuine changes from artifacts, even in challenging high-density urban environments.

\section{Discussion and analysis} \label{sec_discussion}
\subsection{Ablation study}
\begin{figure*}[h]
    \centering
    \includegraphics[width=\linewidth]{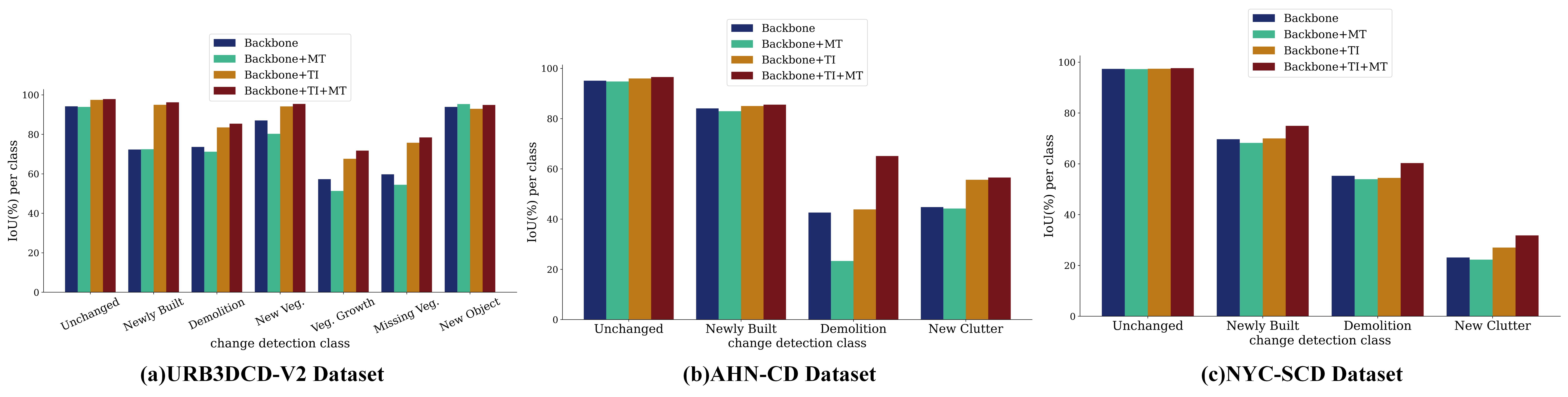}
    \caption{Quantitative results of ablation experiments of the proposed method on different airborne datasets.}
    \label{fig_ablation}
\end{figure*}

In this section, ablation experiments are conducted to evaluate the effectiveness of different mechanisms in the proposed framework, with experiments performed on three airborne change detection datasets. The proposed framework uses a cross-temporal transformer as the backbone, which is further enhanced by the addition of temporal indicators encoding (TI) and multi-task mechanism (MT). The performance of the model with the integration of the TI module and the MT module is tested to explore the effectiveness of these individual components, and the results are shown in Fig.\ref{fig_ablation}.

\begin{table*}[]
\caption{uantitative results of transfer learning experiments based on different datasets. \label{tab:transfer}}
\centering
\scalebox{0.9}{
\begin{tabular}{lllllll}
\hline      
\multirow{2}{*}{Train Set} & \multirow{2}{*}{Test Set} & \multicolumn{1}{l}{\multirow{2}{*}{mIoU(\%)}} & \multicolumn{4}{c}{IoU(\%)}                                                                                                         \\ \cline{4-7} 
                           &                           & \multicolumn{1}{l}{}                         & \multicolumn{1}{l}{Unchanged} & \multicolumn{1}{l}{Newly Built} & \multicolumn{1}{l}{Demolition} & \multicolumn{1}{l}{New Clutter} \\ \hline
NYC                        & \multicolumn{1}{l|}{NYC}  & 66.17                                        & 97.65                         & 74.93                           & 60.27                          & 31.82                           \\
URB3DCD-V2                 & \multicolumn{1}{l|}{NYC}  & 28.92                                        & 81.75                         & 23.72                           & 6.22                           & 3.98                            \\
ANH                        & \multicolumn{1}{l|}{NYC}  & 41.36                                        & 95.54                         & 21.49                           & 28.22                          & 20.17                           \\ \hline
AHN                        & \multicolumn{1}{l|}{AHN}  & 75.94                                        & 96.56                         & 85.55                           & 65.1                           & 56.55                           \\
URB3DCD-V2                 & \multicolumn{1}{l|}{AHN}  & 18.10                                        & 41.06                         & 22.43                           & 2.16                           & 6.74                            \\
NYC                        & \multicolumn{1}{l|}{AHN}  & 53.70                                        & 95.75                         & 78.63                           & 13.74                          & 26.66                          
 \\ \hline       
\end{tabular}} 
\end{table*}
\begin{table}[h]
\caption{The class distribution of the 4-class change detection in different airborne datasets. \label{tab:class}}
\centering
\scalebox{0.85}{
\begin{tabular}{lllllll}
\hline      
\multirow{2}{*}{Dataset}                      & \multirow{2}{*}{Subset} & \multicolumn{4}{l}{Percentage of the different classes(\%)} \\ \cline{3-6} 
                                              &                         & Unchanged   & Newly built   & Demolition   & New Clutter   \\ \hline
\multicolumn{1}{l|}{\multirow{3}{*}{AHN-CD}}  & Train                   & 92.03       & 6.02          & 0.93         & 1.02          \\
\multicolumn{1}{l|}{}                         & Val                     & 91.56       & 6.82          & 0.23         & 1.38          \\
\multicolumn{1}{l|}{}                         & Test                    & 81.79       & 5.04          & 1.99         & 11.18         \\ \hline
\multicolumn{1}{l|}{\multirow{3}{*}{NYC-SCD}} & Train                   & 96.56       & 2.33          & 0.63         & 0.48          \\
\multicolumn{1}{l|}{}                         & Val                     & 96.44       & 2.44          & 0.54         & 0.59          \\
\multicolumn{1}{l|}{}                         & Test                    & 97.15       & 1.84          & 0.53         & 0.47         
   \\ \hline       
\end{tabular}} 
\end{table}
When the cross-temporal transformer serves as the backbone, adding the temporal indicators encoding (TI) improves the accuracy significantly on the Urb3DCD-V2 and AHN-CD datasets. This suggests that when the network applies local attention, it can better distinguish the differences within the patch and, as a result, learn more accurate features, thus enhancing change detection performance. However, when only the multi-task mechanism (MT) is added to the backbone, the accuracy decreases. This is because the points from different epochs within the same patch are not effectively distinguished when extracting features. The addition of multiple supervision signals causes confusion in the network’s features, leading to a drop in accuracy. In contrast, when both the multi-task mechanism and temporal indicators encoding are applied simultaneously to enhance the backbone, multi-class change detection accuracy improves across all three datasets.

From the results shown in Fig.\ref{fig_ablation} (a) and (c), it is evident that the categories of vegetation growth, building demolition, and vegetation removal are more sensitive to different modules. These categories, which are less frequently supervised, require rich semantic and change features for accurate detection. Effectively extracting both semantic and change features can significantly improve the accuracy of detecting smaller categories.

In conclusion, compared to binary change detection, which only needs to identify geometric changes without distinguishing between multiple categories, multi-class change detection requires the accurate extraction of differential features while capturing the semantic differences across different epochs.

\subsection{Generalization test}
This section explores the generalization capability and adaptability of change detection models across different datasets. Training is performed on Urb3DCD-V2, AHN-CD, and NYC-SCD, and testing is conducted on AHN-CD and NYC-SCD, respectively. For the simulated dataset, its seven categories are merged into four for training. The results of training and testing on different datasets are presented in the Tab.\ref{tab:transfer}.

Simulated datasets have a low annotation cost, making them an ideal choice for training models that can later be transferred to real-world scenarios for change detection. However, as shown by the results in Tab.\ref{tab:transfer} , directly applying models trained on simulated datasets to real-world data reveals significant differences between the datasets. These differences arise not only from the way simulated data are generated compared to real ALS data, but also from the varying distribution of change categories. Tab.\ref{tab:class} summarizes the proportions of points in four change detection categories across different datasets. In particular, there is a significant discrepancy in the sample distributions between the simulated data set and the real world data sets (AHN-CD and NYC-SCD). For instance, in the simulated dataset, the proportions of newlt built and demolition are nearly equal, whereas in real-world scenarios, the proportion of demolition is much smaller.

Training in individual real-world datasets and testing on unseen real-world scenarios yields better results compared to models trained solely on simulated datasets. This is due to the shared characteristics among categories in real-world datasets. Additionally, the diversity of a dataset plays a crucial role in the accuracy of transfer learning. For example, in experiments on NYC-SCD, the model trained on NYC-SCD demonstrates high performance for the new building category in AHN-CD because NYC-SCD includes diverse examples of new construction, encompassing most types found in real-world scenarios. However, the results for the demolition category is less effective. This is partly because demolition constitutes a smaller proportion in NYC-SCD and partly due to differences in urban change patterns between the different datasets. In addition, NYC-SCD predominantly experiences growth, with a greater variety of new buildings.

Combining experimental results and analysis, if an existing dataset provides diverse and comprehensive labeled data, and the transfer process incorporates a small amount of additional supervision, the generalization capability of change detection models for real-world scenarios can be significantly improved. Future work could further explore more efficient transfer methods, such as unsupervised domain adaptation or few-shot learning, to address the challenges of dataset differences effectively.

\section{Conclusion} \label{sec_conclusion}

In summary, this work addresses critical challenges in urban 3D semantic change detection, including limited feature association across different epochs, class imbalances, and insufficient dataset diversity. We proposed the Multi-Task Enhanced Cross-Temporal Point Transformer (ME-CPT), which integrates spatial-temporal modeling, cross-temporal attention mechanisms, and multi-task training strategies to improve feature extraction and alignment, ensuring robust detection of multi-class changes. Additionally, the release of a 22.5 $km^2$ 3D semantic change detection dataset provides a valuable resource for benchmarking and advancing research in this field. 
3D change detection based on multi-temporal point clouds enables the most detailed urban change detection results, supporting urban management and 3D map updates. 
In future work, we will explore more lightweight networks to enable detection of a greater number of change categories, such as vegetation removal and vegetation growth, in real-world scenarios. This will help meet the needs of applications such as urban asset management and vegetation surveys.

\setlength{\bibsep}{2pt} 
\renewcommand{\bibfont}{\small}
\bibliography{refs} 
\bibliographystyle{IEEEtranN}


\end{document}